\documentclass[a4paper,fleqn]{cas-dc}

\ExplSyntaxOn
\exp_args:NNno \exp_args:Nno \use:n { \cs_gset:Npn \__make_fig_caption:nn #1#2 }
  {
    \exp_after:wN \use_ii_i:nn \exp_after:wN
      { \__make_fig_caption:nn {#1} {#2} }
      { \dim_set:Nn \l_fig_width_dim \linewidth }
  }
\exp_args:NNno \exp_args:Nno \use:n { \cs_gset:Npn \__make_tbl_caption:nn #1#2 }
  {
    \exp_after:wN \use_ii_i:nn \exp_after:wN
      { \__make_tbl_caption:nn {#1} {#2} }
      { \dim_set:Nn \l_tbl_width_dim \linewidth }
  }
\ExplSyntaxOff

\usepackage{amsmath, amssymb, amsfonts}
\usepackage{graphicx}
\usepackage{booktabs}
\usepackage{algorithm, algorithmic}
\usepackage{enumitem}
\usepackage{caption}
\usepackage{array}
\usepackage{stfloats}
\usepackage{url}
\usepackage{verbatim}
\usepackage{cite}
\usepackage{color}
\usepackage{tabularx}
\usepackage{multirow}
\usepackage[numbers]{natbib}
\usepackage{hyperref}
\usepackage{makecell}
\usepackage{threeparttable}

\newcommand{\ie}{\emph{i.e.},\xspace}
\newcommand{\eg}{\emph{e.g.},\xspace}

\usepackage{etoolbox}
\usepackage{flushend}
\usepackage{lettrine}
\def\tsc#1{\csdef{#1}{\textsc{\lowercase{#1}}\xspace}}
\tsc{WGM}
\tsc{QE}
\tsc{EP}
\tsc{PMS}
\tsc{BEC}
\tsc{DE}

\begin{document}
\let\WriteBookmarks\relax
\def\floatpagepagefraction{1}
\def\textpagefraction{.001}

\shorttitle{A comprehensive survey on deep learning techniques in educational data mining}
\shortauthors{Y. Lin et~al.}

\title [mode = title]{A comprehensive survey on deep learning techniques in educational data mining}

%
%
%
\author[1,2]{Yuanguo Lin}[style=chinese]
\fnmark[1]
\ead{xdlyg@jmu.edu.cn}

\author[3]{Hong Chen}[style=chinese]
\fnmark[1]
\ead{zoharchen30@gmail.com}

\author[4]{Wei Xia}[style=chinese]
\ead{xiawei24@huawei.com}

\author[2]{Fan Lin}[style=chinese]
\cormark[1]
\ead{iamafan@xmu.edu.cn}


\author[1]{Zongyue Wang}[style=chinese]
\cormark[1]
\ead{wangzongyue@jmu.edu.cn}

\author[5]{Yong Liu}
\ead{stephenliu@ntu.edu.sg}

\address[1]{School of Computer Engineering, Jimei University, Xiamen, China}
\address[2]{School of Informatics, Xiamen University, Xiamen, China}
\address[3]{Pingtan Research Institute of Xiamen University, Fuzhou, China}
\address[4]{Huawei Noah’s Ark Lab, Shenzhen, China}
\address[5]{Huawei Noah’s Ark Lab, Singapore}

\fntext[1]{Co-first authors}
\cortext[1]{Corresponding author}

\newcommand{\xwei}[1]{{\bf \color{cyan} [xwei says ``#1'']}}

\begin{abstract}
Educational Data Mining (EDM) has emerged as a vital field of research, which harnesses the power of computational techniques to analyze educational data. With the increasing complexity and diversity of educational data, Deep Learning techniques have shown significant advantages in addressing the challenges associated with analyzing and modeling this data. This survey aims to systematically review the state-of-the-art in EDM with Deep Learning. We begin by providing a brief introduction to EDM and Deep Learning, highlighting their relevance in the context of modern education. Next, we present a detailed review of Deep Learning techniques applied in four typical educational scenarios, including knowledge tracing, student behavior detection, performance prediction, and personalized recommendation. Furthermore, a comprehensive overview of public datasets and processing tools for EDM is provided. We then analyze the practical challenges in EDM and propose targeted solutions. Finally, we point out emerging trends and future directions in this research area.
\end{abstract}

%

\begin{keywords}
Educational data mining \sep Deep learning \sep Reinforcement learning \sep Educational datasets
\end{keywords}

\maketitle

\section{Introduction}

Deep Learning has experienced remarkable advancements in recent years, revolutionizing diverse domains, including education. Deep Learning, a form of machine learning, relies on artificial neural networks to facilitate the discovery of hierarchical features, which enhances the ability to recognize patterns~\cite{lecun2015deep}. In contrast to conventional machine learning approaches that require manual feature engineering, Deep Learning allows a machine to automatically discover intricate structures in large data by using multiple layers of abstraction. This layered feature learning process enables Deep Learning models to learn complicated patterns in data and achieve state-of-the-art performance across domains such as speech recognition~\cite{bai2021speaker}, image classification~\cite{touvron2022resmlp}, and natural language processing~\cite{roh2021unsupervised}. In general, three primary categories of algorithms exist in the domain of Deep Learning: supervised learning, unsupervised learning, and reinforcement learning~\cite{mathew2021deep}. Applying Deep Learning algorithms suitable for different application scenarios~\cite{huang2023examining,liu2022automated} can greatly increase their performance.
With increasing computing power, Deep Learning has achieved major breakthroughs and outstanding results in many fields.

Education fields have been revolutionized by traditional machine learning and Deep Learning algorithms, and exploring previous research is key to understanding its applications. There are two main terms which are currently used in education, Educational Data Mining (EDM) and Learning Analytics (LA) \cite{romero2020educational}. These two fields are typically interdisciplinary, including but not limited to information retrieval, data analysis, psycho-pedagogy, cognitive psychology, etc. EDM encompasses computerized techniques and tools that facilitate the automated identification and extraction of meaningful patterns and valuable information from extensive datasets obtained within educational settings \cite{kumar2017data,du2020educational,liu2024xes3g5m}. On the other hand, LA involves the systematic gathering, examination, and presentation of data pertaining to learners and their learning environments \cite{siemens2012learning}.

In fact, researchers have also utilized diverse traditional machine learning algorithms~\cite{chen2024applying} for EDM across varied educational contexts~\cite{li2022dkt}. For instance, the TLBO-ML model~\cite{arashpour2023predicting} constructed by Artificial Neural Network (ANN) and Support Vector Machine (SVM) can be used to predict grades of students in final exam. 
However, traditional machine learning algorithms come with certain limitations. These conventional models often rely on manual feature engineering, necessitating expert knowledge in the field to design effective features. This process can be laborious and time-consuming. Besides, handling the complex and high-dimensional data, such as natural language text and multimedia content that are prevalent in educational settings, can pose challenges for traditional machine learning methods~\cite{morciano2024use}.

Integrating Deep Learning into educational scenarios is driven by the aspiration to harness the potential of Artificial Intelligence and machine learning, thereby enriching the teaching and learning experience \cite{lecun2015deep, hernandez2019systematic}. Deep Learning models possess remarkable capabilities, enabling them to effectively process and analyze vast amounts of educational data. By uncovering meaningful patterns and making accurate predictions, these models provide valuable insights that can inform and enhance educational practices \cite{song2022survey, sun2019deep, safarov2023deep}. Educators and researchers can leverage these insights to adapt teaching strategies, personalize instruction, and optimize learning outcomes. By harnessing the power of Deep Learning, the educational community can unlock new opportunities for improved efficiency, effectiveness, and adaptability in education through intelligent data processing and adaptive methodologies \cite{perrotta2020deep}. 

In addition, Deep Learning has also made significant achievements in completing specific scenario tasks in EDM. Currently, EDM scenarios are often divided into: knowledge tracing, student behavior detection, performance prediction, and personalized recommendation. Each subfield has distinct specific data input patterns and task requirements. Deep Learning-based knowledge tracing algorithms~\cite{song2022survey} can be classified into Deep Knowledge Tracking (DKT) and its variants, \eg DKT based on Memory networks, Attention mechanisms~\cite{zhan2024contrastive} and Graph structures~\cite{cui2024dgekt}. Complex neural network models can be applied to Student Dropout Prediction (SDP), which is a task of student behavior detection~\cite{prenkaj2020survey}. Similarly, neural networks can also provide quite reliable accuracy in the EDM scenario of performance prediction~\cite{rastrollo2020analyzing}. In the field of personalized recommendation~\cite{huang2019exploring}, hybrid techniques-based recommendation algorithms may be gradually taking the dominant position, but the emerging privacy issues also need close attention~\cite{khanal2020systematic}.

Deep Learning techniques offer numerous advantages over traditional machine learning methods when applied in educational scenarios~\cite{huang2024response,li2024quantification}. One notable advantage is the automatic learning of hierarchical representations from raw data, eliminating the need for manual feature engineering. This characteristic makes Deep Learning models highly suitable for analyzing various types of educational data, including student performance data, educational videos, and text-based learning materials. The capability of Deep Learning models to detect complex connections and intricate patterns within the data leads to more accurate predictions and personalized recommendations. A telling difference is shown by a research \cite{hernandez2019systematic} that approximate 67\% of papers report Deep Learning demonstrated superior performance compared to the ''traditional'' machine learning baselines in all conducted experiments.

\textbf{Data Collection Methodology.} Followings are rules we applied to include or exclude papers:
\begin{itemize}
    \item Search terms: Deep Learning and Educational Data Mining are the two keywords mainly involved in our survey. In order to access more related publications, we also used specific educational scenarios as our search terms (\eg knowledge tracing, performance prediction etc.).
    \item Search Sources: We searched for articles containing the above keywords through Google Scholar and downloaded the articles that matched the requirements from the corresponding major databases.
    \item The articles we have studied include only high-level publications from international conferences and top journals based on the application of Deep Learning to educational scenarios.
\end{itemize}

\begin{figure}
    \centering
    \includegraphics[width=0.45\textwidth]{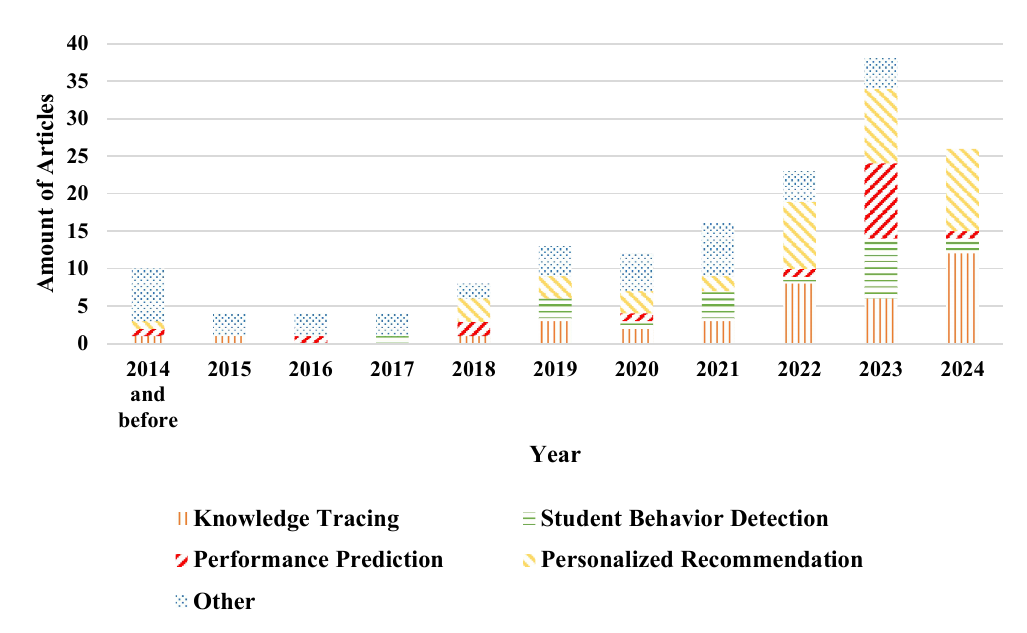}
    \caption{Distribution of year-wise publications (until May 2024) according to different educational scenarios.}
    \label{fig:dis}
\end{figure}

We plotted the collected papers according to their publication years and corresponding application scenarios as shown in Fig. \ref{fig:dis}, and to ensure the frontiers of research, it can be seen that EDM based on Deep Learning has emerged since 2018.
\par 

\textbf{Related Work.} Ahmad et al.~\cite{ahmad2023data} examine the application of artificial intelligence in EDM, specifically focusing on the use of standard machine learning techniques for student assessment and prediction.
Besides, Sun et al.~\cite{sun2021brief} explore the application of Deep Learning in MOOCs for dropout prediction, focusing on the trends in using Deep Learning techniques, including feature processing and model design.
Zhang et al. \cite{zhang2018survey} also provide an overview of the educational data field, covering a range of classic Deep Learning models such as Stacked Auto-Encoder (SAE), Deep Belief Network (DBN), and various Deep Neural Networks (DNN). It further investigates distinct Deep Learning techniques tailored for different types of educational data, including large-scale, heterogeneous, real-time, and low-quality data.
\textbf{Our Contribution.} This survey encompasses a diverse collection of state-of-the-art studies that categorize Deep Learning algorithms based on their application in educational settings. The articles are classified into three types: supervised learning, unsupervised learning, and reinforcement learning. Furthermore, the articles are further categorized based on the specific educational scenarios where Deep Learning techniques are employed. These scenarios encompass knowledge tracing, student behavior detection, performance prediction, and personalized recommendation. By adopting this systematic approach, we provides a structured and in-depth analysis of the wide range of research conducted on the intersection of Deep Learning and education.

This survey not only summarizes the latest Deep Learning algorithms, but also includes a discussion of relevant datasets and data tools. By including a comprehensive analysis of datasets and data tools, the main objective of this survey is to offer the reader a thorough understanding of the considerations and challenges when implementing Deep Learning solutions in EDM. To this end, we perform a comprehensive analysis of the practical challenges (\eg data quality and integration, evaluation and validation, scalability and real-time processing, fairness and privacy) in EDM, and present targeted strategies to address these challenges. Furthermore, we put forward some promising areas for further research in Deep Learning-based EDM. The potential directions contain learning analysis and intervention, social network analysis and collaboration, explainable AI in EDM, Large Language Models for education, multimodal learning analytics.

In summary, the application of Deep Learning to education has the potential to revolutionize the way traditional teaching and learning. By leveraging the benefits of Deep Learning, educators can gain valuable insights, make data-based decisions, and create personalized learning experiences. This survey aims to provide an integrate overview of Deep Learning applications in education, covering various algorithms and educational scenarios while considering the available datasets and data tools.

\section{Methodology}
In recent years, Deep Learning has emerged as a state-of-the-art technology that can be applied to various fields.

The ability of neural networks to extract higher-level abstract features by learning the features of data has made Deep Learning a highly successful method.

Deep Learning is a technology derived from machine learning. It mimics the structure and working mode of the human brain neural network. It realizes the recognition and classification tasks through model training. Compared to conventional machine learning algorithms, Deep Learning is able to handle more complicated tasks and datasets which brings a higher performance and generalization abilities.

The development of Deep Learning has gone through many stages, from the initial DBN~\citep{hinton2006fast} to the subsequent emerging Multilayer Perceptron (MLP), Convolutional Neural Network (CNN), Recurrent Neural Networks (RNN), etc, which have led to the Deep Learning that we care about and use today. With the progress of Deep Learning algorithms and computing power, Deep Learning has been widely applied to areas such as image recognition, speech recognition, and so on. In this survey, we will mainly focus on the Deep Learning algorithms that contribute to educational scenarios like knowledge tracing, student behavior detection and course recommendation.

We divide those models into three parts according to the current mainstream classification method: supervised learning, unsupervised learning and reinforcement learning~\cite{alzubaidi2021review}.

\subsection{Supervised Learning}
\par Supervised learning usually refers to those models which apply labeled data for training process. These data can be used to train supervised learning models to establish mappings between inputs and outputs and predict the result of the new unlabeled data.
\par CNN stands as a representative example of Deep Learning models which be widely used in computer vision and image process fields. The theory of CNN is to extract feature and reduce the dimension of the image through convolution and pooling operations. Finally, the high-dimensional image is converted into the one-dimensional vector data which can be used for the classification and regression tasks, \eg the sentiment analysis of courses~\cite{bhanuse2024optimal}.
\par The difference between CNN and other neural networks is the former applies convoluted layer to extract features as the following formula~\cite{alzubaidi2021review}: 
\begin{equation}
    h^k = f(W^k *x+b^k)
    \label{eq:1}
\end{equation}
where $h^k$ refers to the feature map generated by $k_{th}$ convoluted kernel, $W^k$ represents the weight of $k_{th}$ convoluted kernel, $b^k$ is the corresponding bias term and $f(\cdot)$ stands for activation function. CNN has been applied for knowledge tracing. A model called Deep Knowledge Tracing based on Spatial and Temporal Deep Representation Learning for Learning Performance Prediction (DKT-STDRL) proposed by Lyu et al.~\cite{lyu2022deep}. In this model, CNN plays the role of extracting the spatial features information from students’ exercise sequences.
\par RNN is primarily employed for handling sequential data, such as text and speech, in various applications. The neural nodes of RNN are able to receive the former status information to realize the memorability. The special property allows RNN to be applied to knowledge tracing, student behaviors detection and similar educational scenarios. The following is a basic RNN equation for it to be able to implement cycling:
\begin{equation}
    h_t=\sigma(W_{ih}x_t+W_{hh}h_{t-1}+b_h),
    \label{eq:2}
\end{equation}
where $x_t$ refers to the input vector, The $h_t$ is the hidden state vector, $W_{ih}$ and $W_{hh}$ are the weight matrices connecting the input to the hidden state and the hidden state to the hidden state, $b_h$ is the bias vector, and $\sigma$ is a nonlinear activation function. 
\par
\begin{figure}
    \centering
    \includegraphics[width=0.4\textwidth]{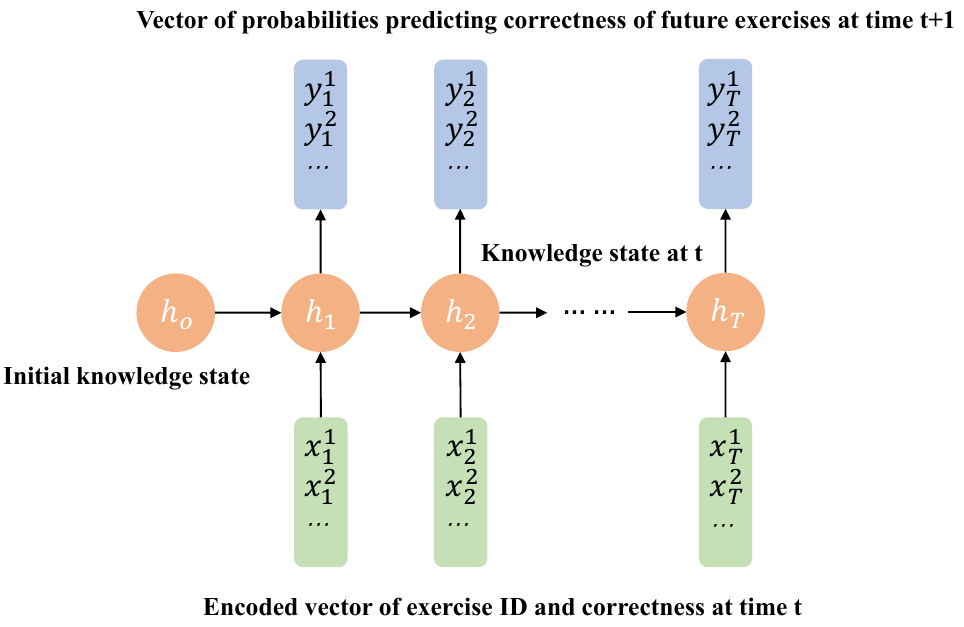}
    \caption{The process of Recurrent Neural Network~\cite{piech2015deep}, taking knowledge tracing as an example.}
    \label{fig:1}
\end{figure}

 Fig. \ref{fig:1} simply shows the work process of RNN. Based on that, a Deep Knowledge Tracing model (DKT) implemented with RNN was proposed by Piech et al.~\cite{piech2015deep} This DKT model employs a significant number of artificial neuron vectors to represent potential knowledge states and temporal dynamics.
\par Long Short-Term Memory (LSTM)~\cite{hochreiter1997long} is a special RNN that contains three gate units and a memory cell. The memory and processing of sequential data are realized through the control of information flow by gated units. Importantly, the design of LSTM, with its gates and cell states, addresses the gradient vanishing problem more effectively than conventional RNNs. This is achieved by allowing gradients to flow through the network with less restriction, thereby preserving the necessary information over longer sequences. The forgetting gate is a key component in LSTM to control whether the previous moment's memory cell state is forgotten, shown in Eq. \ref{eq:3}.
\begin{equation}
    f_t=\sigma(W_f\cdot[h_{t-1},x_t]+b_f),
    \label{eq:3}
\end{equation}
where $f_t$ denotes the state of the forgotten gate, $W_f$ is the weight matrix, $[h_{t-1},x_t]$ denotes the connection between the output from the previous moment and the input from the current moment, $b_f$ is the bias vector, and $\sigma$ is the sigmoid function.
\par Graph Neural Network (GNN)~\cite{wu2022graph} is a Deep Learning model used to process graph structure data (\eg Knowledge Graph). Building on this, GNNs are particularly adept at discerning intricate patterns and dependencies within the graph, enabling more precise and context-aware course recommendations~\cite{wang2021dskreg}. Take Recurrent GNN as an example, the recurrently update formula of a node's hidden state can be defined by~\cite{wu2020comprehensive}:
\begin{equation}
    \mathbf{h}^{(t)}_v=\sum_{u \in N(v)}f(\mathbf{X}_v,\mathbf{X}_{(v,u)}^\mathbf{e},\mathbf{X}_u,\mathbf{h}_u^{(t-1)}),
    \label{eq:4}
\end{equation}
where $f$ represents the parametric function and  $\mathbf{h}_u^{(t-1)}$ is initialized randomly. $\mathbf{X}_v$ refers to feature vector of node $v$, $\mathbf{X}_{(v,u)}^\mathbf{e}$ denotes edge vector of node $(v,u)$.
\par Because of its excellent graph structure processing capability, GNN has been used by the Knowledge Augmented User Representation (KAUR)~\cite{ma2023enhancing} model to obtain the initial representation of a specific node within the Collaborative Knowledge Graph (CKG), for the message aggregation and propagation.


\par Attention is mainly used to enhance the model's focus and understanding of key information. The core idea is to assign weights to different parts of the input according to task requirements, allowing the model to selectively focus on specific information~\cite{lin2021adaptive}. The typical Self-Attention mechanism can be expressed by the following formula:
\begin{equation}
    \text{Attention}(Q,K,V)=\text{Softmax}(\frac{QK^T}{\sqrt{d_k}})V,
\end{equation}
where $Q$ means a matrix packaged from queries, $K$ refers to the key, $V$ means value, and $d_k$ is the key dimension. By calculating the similarity between the query and the key and normalizing it, the Attention weight distribution is calculated and ultimately used to weight the sum values to obtain the final representation. Attention-based networks \cite{vaswani2017attention}, dispensing with recurrence and convolutions operation. It can be trained in parallel and achieve better performance.

\subsection{Unsupervised Learning}
\par Unsupervised learning's aptitude for handling unlabeled data empowers it to uncover latent patterns in student behavior and learning processes. While algorithms like Generative Adversarial Networks and Autoencoders are not commonly employed in education, they still hold promise for enhancing educational insights by extracting valuable information from unlabeled data.
\par Generative Adversarial Network (GAN) is a combination of generator neural network and discrimination neural network. The target of GAN is to learn and generate samples that resemble real data, and with a high degree of diversity. The loss function of the discriminator is listed as follows: 
\begin{equation}
    \begin{split}
        L_D&=-[\log(D(x))+\log(1-D(G(z)))] \\
        L_G&=-\log(D(G(z))),
    \end{split}
\end{equation}
where $x$ denotes real samples, $D(x)$ is the output of discriminator towards real sample, $G(z)$ is the sample generated by generator and $1-D(G(z))$ is the output of discriminator towards generative samples. Similarly, the lost function of the generator is as follows, $z$ is a random sample get from noise distribution, $G(z)$ is the sample generated by the generator, and $D(x)$ is the output of the discriminator towards sample $x$. 
\par The most common use of GAN in education scenarios is a personalized recommendation because of its ability to generate samples. For instance, a Recurrent Generative Adversarial Network (RecGAN) that leverages tailored GRU to obtain latent features of users and items from short/long-term temporal profiles was introduced by Bharadhwaj et al.~\cite{bharadhwaj2018recgan}. By evaluation, this method has improved the relevance of recommended items.


\par AutoEncoders (AEs) are neural networks used for unsupervised learning, focusing on encoding and then reconstructing input data. The key formula involves two parts~\cite{zhai2018autoencoder}:
\begin{equation}
    \begin{split}
        &\text{Encoder:}z=f(x)\\
        &\text{Decoder:}\hat{x}=g(z),
        \label{eq:8}
        \end{split}
\end{equation}
where $x$ is the input, $z$ represents the encoded representation, $\hat{x}$ stands for the reconstructed output, $f(\cdot)$ represents the encoding function, and $g(\cdot)$ is the decoding function. The aim is to minimize the difference between $x$ and $\hat{x}$, enhancing data compression and feature learning capabilities.


\subsection{Reinforcement Learning}
\par In contrast to supervised and unsupervised learning, the training data used in reinforcement learning is traditionally generated through the interaction of an agent with its environment. Therefore, reinforcement learning is suitable for personalized recommendation, which can be considered as an intelligent tutoring task. It allows the model to interact dynamically and continuously adjust recommendation strategies to maximize long-term rewards. It can optimize recommendation effectiveness based on real-time user feedback.
\par In reinforcement learning~\cite{zhang2022self}, an agent takes a specific action by observing the state of the environment and evaluates its behavior based on the reward or punishment given by the environment. The goal of the agent is to optimize the cumulative reward. Generally, the reinforcement learning algorithms can be divided into the following three styles.

\subsubsection{Value Function Approach}
Value function approach refers to the algorithm's ability to achieve the global optimal payoff by obtaining the best action. That is, the optimal gain is produced through the optimal action $a^*$ under the optimal strategy $\pi^*$. This strategy can be represented by the Bellman optimality equation:
\begin{equation}
    \begin{split}
        v^*_\pi(s)\ = \ &\underset{a}{\text{max}}\ \mathbb{E}[R_{t+1}+\gamma v^*_\pi (S_t+1)|S_t = s,A_t = a]\\
        = \ &\underset{a}{\text{max}}\sum_{s',r}p(s',r|s,a)[r+\gamma v^*_\pi (s')],
    \end{split}
\end{equation}
where $\mathbb{E[\cdot]}$ denotes the expectation of the reward $R_{t+1}$ and value function $\gamma v^*_\pi (S_t+1)$ for the next state in the case of the current state $s$ and taking action $a$.
\par The ability of the value function to take into account long-term benefits allows the model to make more informed decisions, not just localized immediate rewards. However, the solution of certain value function methods may take a longer time to reach convergence. Especially in complex environments or large-scale problems, more iterations and samples may be required to obtain accurate value function estimates and optimal strategies.

\subsubsection{Policy Search Method}
The policy search method maximizes the expected return in a direct optimization of strategies that are influenced by a set of policy parameters $\theta_t$. As an example, the gradient-based policy search method uses the Gradient Ascent method, which maximizes the strategy performance $J$ with respect to the parameter $\theta$ by iteratively updating the strategy parameters, and the equation can be simply expressed as:
\begin{equation}
    \theta_{t+1}\ = \ \theta_t + \alpha \nabla J(\theta),
\end{equation}
where $\theta_{t+1}$ refers to the parameter of policy at time $t+1$, $\nabla J(\theta)$ is the gradient of $\theta_t$-based policy's performance. $\alpha$ is a learning rate that controls the step size of each parameter update.
\par In addition to gradient-based policy search algorithms, there is a Monte Carlo policy gradient-based method called REINFORCE for optimizing policies in reinforcement learning. It estimates the policy gradient by Monte Carlo sampling and uses the gradient ascent method to update the policy parameters. Specifically, the REINFORCE algorithm can be implemented by the following equation:
\begin{equation}
    \begin{split}
        \nabla J(\theta) \ &\propto \ \sum_s \mu(s) \sum_a  \nabla_\theta \pi (a|s,\theta) q_\pi (s,a)\\
        &\doteq \ \mathbb{E}_\pi \left [\sum_a \nabla_\theta \pi (a|S_t,\theta) q_\pi (S_t,a)\right ],
    \end{split}
\end{equation}
where $\propto$ denotes "be proportional to", $\mu(s)$ is called "on-policy" under the policy $\pi$. $q_\pi (s,a)$ refers to the value function of policy $\pi$ choosing action $a$ in state $s$.

\subsubsection{Actor-Critic Algorithm}
The Actor-Critic (AC) algorithm is a reinforcement learning algorithm for solving the problem of learning optimal policies in unknown environments. The core idea of the Actor-Critic algorithm is to guide Actor's policy improvement through the estimation of value functions provided by the Critic, which in turn estimates the value of a state or state action pair based on the current policy and environment interaction data.
\par Take one-step Actor-Critic algorithm as an example, the equation of $\theta$ update can be expressed as:
\begin{equation}
    \resizebox{0.88\hsize}{!}{$
        \theta_{t+1} \doteq \theta_t + \alpha(R_{t+1} + \gamma \hat{v}(S_{t+1},w)-\hat{v}(S_t,w))\frac{\nabla_\theta \pi(A_t|S_t,\theta_t)}{\pi(A_t|S_t,\theta_t)},
    $}
\end{equation}
where $R_{t+1}$ denotes the reward at time $t+1$, $\gamma$ is the discount factor that controls the importance of future rewards, and $\hat{v}(S_{t},w)$ refers to the state value function learned by the Critic and will be used as baseline.

\section{Educational Scenarios and Corresponding Algorithms}
To provide a comprehensive overview of the literature on Deep Learning for different application scenarios, we have included information on the algorithms and models used in each paper for four educational scenarios in Table \ref{tab2}. This table also lists the evaluation metrics and datasets employed in each study.
\begin{table*}[htbp]
\small
  \centering
  \caption{Overview of Deep Learning Algorithms for Different Educational Scenarios.}
   \resizebox{\linewidth}{!}{
    \begin{tabular}{llllllr}
    \toprule
    \textbf{Scenario} & \textbf{Model} & \textbf{Algorithm Classification} & \textbf{Method} & \textbf{Evaluation Metric} & \textbf{Dataset} & \textbf{Year}\\
    \midrule
    \multirow{23}[0]{*}{Knowledge Tracing} & GFLDKT~~\cite{zhao2023novel} & \multirow{18}[0]{*}{Supervised Learning} & LSTM  & AUC/ACC & ASSISTments15/17/JunyiAcademy&2023  \\
          & LFBKT~~\cite{chen2022knowledge} &       & LSTM  & AUC/ACC & ASSISTment12&2022 \\
          & DFKT~~\cite{nagatani2019augmenting}&       &LSTM &AUC/ACC/MAP&ASSISTment12&2019\\
          & PGN~~\cite{li2023plastic}   &       & RNN    & AUC/ACC/RMSE & ASSISTment15/17/Statics2011&2023\\
          & DKT~~\cite{piech2015deep} &       & RNN & AUC & ASSISTment09/10&2015\\
          & DKT-STDRL~~\cite{lyu2022deep} &       & CNN   & AUC/ACC/RMSE/$r^2$ & ASSISTment09/15/Statics2011 &2022\\
          & HHSKT~~\cite{ni2023hhskt} &       & GNN & AUC/ACC & ASSISTment09/17/Junyi15&2023\\
          & TSKT~~\cite{yang2024heterogeneous} &       & GNN & AUC/ACC & ASSISTments09/15/Eanalyst&2024\\
          & GIKT~~\cite{yang2021gikt} &       & GCN & AUC     & ASSISTment09/12/EdNet&2021\\
          & FCIKT~~\cite{xu2024modeling} &       & GCN/Attention & AUC & ASSISTments09/12/EdNet&2024\\
          & MRTKT~~\cite{cui2023fine} &       & Attention    & AUC/ACC & ASSISTment09/10/12/13&2023\\
          & KTMFF~~\cite{xiao2023knowledge} &       & Attention & AUC   & ASSISTment09/15/Statics2011&2023\\
          & AKT~~\cite{ghosh2020context}&         &Attention & AUC &ASSISTment09/10/12/13/KDD Cup2012&2020\\
          &SAKT~~\cite{pandey2019self}&   &Attention &AUC &ASSISTment09/15/Statis2011&2019\\
          &TCKT~~\cite{huang2024learning}&   &Attention &AUC/ACC/RMSE &ASSIST2012/ASSISTChall/EdNet&2024\\
          &ARAIKT~~\cite{wong2024architectural}&   &Attention &AUC/ACC &ASSISTments09/17&2024\\
          &ELAKT~~\cite{pu2024elakt}&   &Transformer &AUC/ACC/MAE/RMSE &ASSISTments09-10/15/17/EdNet&2024\\
          &ENAS-KT~~\cite{yang2024evolutionary}&   &Transformer &AUC/ACC/RMSE &EdNet/RAIEd2020&2024\\ 
          \cmidrule{2-7}          
          & AdaptKT~~\cite{cheng2022adaptkt} & \multirow{1}[0]{*}{Unsupervised Learning} & Autoencoder & AUC   & zx.math/ax.physics&2022\\
\cmidrule{2-7}          & KTM~~\cite{cai2019learning} & \multirow{4}[0]{*}{Reinforcement Learning} & AC & AUC   & ASSISTment09/10&2019\\
&RL-KTNet~~\cite{ding2020automatic}&&AC&AUC/$r^2$&ASSISTment09/10/KKD2010&2020\\
&IEKT~~\cite{long2021tracing} &&Policy Gradient&AUC/ACC&ASSISTment09/12/EdNet/Junyi&2021\\
&KADT~~\cite{abdelrahman2023learning}&&DPG&AUC&ASSITment09/IMDB/MovieLens/CIFAR-100&2023\\
    \midrule
    \multirow{12}[0]{*}{Student Behavior Detection} & VB-DTW~~\cite{wang2023automated} & \multirow{11}[0]{*}{Supervised Learning} & CNN   & ACC/VSE & SCB-13&2023\\
          & DCNN~~\cite{hdioud2023facial}  &       & CNN   & Recall/Precision/F1 & FER2013&2023\\
          & ATGCN~~\cite{qiu2023ld} &       & CNN   & ACC   & N/A&2023\\
          & Faster R-CNN~~\cite{komagal2023ptz} &       & CNN   & ACC/Precision/DR/FDR & TCE Classroom&2023\\
          & CFIN~~\cite{feng2019understanding} &     & CNN & AUC/F1 & KDD Cup2015/XuetangX&2019\\
          & EDLN~~\cite{kumar2023ensemble} &     & CNN & AUC/Recall/Precision/F1 & KDD Cup2015 & 2023\\
          & ABDM~~\cite{zhou2023abnormal} & & CNN & ACC & N/A & 2023\\
          & LBDL~~\cite{liu2023learning} & & LSTM & AUC/F1 & MOOCCube & 2023\\
          & CDLSTM~~\cite{alsabhan2023student} & & LSTM & AUC/Recall/Precision/F1 & 7WiseUp & 2023\\
          & ECLSTM~\cite{talebi2024ensemble} &  & LSTM & AUC/Recall/Precision/F1 & KDD Cup2015&2024\\
          & EVA-MLP~~\cite{tan2023emotional} &       & MLP   & Recall/Precision/F1 & Student Journals&2023\\
\cmidrule{2-7}          & FTGAN~~\cite{stenton2021fine} & \multirow{1}[0]{*}{Unsupervised Learning} & GAN   & AUC/ACC/Recall/Precision/F1 & N/A&2021\\
          
    \midrule
    \multirow{12}[0]{*}{Performance Prediction} & DL-MLP~~\cite{chen2023machine} & \multirow{9}[0]{*}{Supervised Learning} & MLP    & ACC & Pennsylvania School Performance Profile&2023\\
          & CRN~~\cite{nayani2023combination}  &       & CNN   & FDR/Sens/FPR/MCC/F1/NPV/FNR & Kaggle&2023\\
          &UPC-Net~\cite{alshamaila2024automatic} & & CNN &Recall/Precision/F1&Self-collected&2024\\
          & SDPNN~~\cite{neha2023deep} &       & DNN   & ACC   & N/A&2023\\
          & MLP-12Ns~~\cite{beckham2023determining} &       & MLP   & RMSE  & Kaggle&2023\\
          &TLBO-ML~~\cite{arashpour2023predicting} & & ANN & ACC/Recall/Precision/FM/F1/MCC & OULAD&2023\\
          & SAPP~~\cite{kukkar2023prediction}  &       & LSTM  & ACC/Recall/Precision/F-measure & OULAD&2023\\
          &GRU-ANOVA~\cite{lakshmi2024effective}  &   & RNN & ACC/Recall/Precision/F1 &Self-collected  &2024\\
          & MSH-MD~~\cite{chen2023prediction} & & Attention & Recall/Precision/F1/MSE/MAE/RSME & Self-Collected & 2023\\
\cmidrule{2-7}          & CGAN~~\cite{sarwat2022predicting}  & Unsupervised Learning & GAN   & AUC   & Self-collected&2022\\
\cmidrule{2-7}          & ADSLS~~\cite{dorcca2013comparing} & \multirow{2}[0]{*}{Reinforcement Learning} & Q-Learning & N/A   & N/A&2013\\
&NCAT~~\cite{zhuang2022fully} &  & Q-Learning & AUC/ACC & ASSISTment09/12/15/KDD Cup2010&2022\\
    \midrule
    \multirow{27}[0]{*}{Peronalized Recommendation} & BERT~~\cite{li2023personalized}  & \multirow{14}[0]{*}{Supervised Learning} & Attention   & Recall/Precision/F1 & MOOCCube&2023\\
          & MCR-C-FGM~~\cite{sakboonyarat2022applied} &       & DNN   & Precision & edX Learning Data&2022\\
          & SODNN~~\cite{safarov2023deep} &       & DNN   & Precision & UBOB&2023\\
          & LSTM-CNN~~\cite{vedavathi2023plrec} &       & CNN   & Recall/Precision/F1 & Kaggle&2023\\
          & CSEM-BCR~~\cite{gao2022online} &       & CNN   & Recall/Precision/MAP/AP & CourseTalk&2022\\
          & ARGE~~\cite{zhao2023agre}  &       & RNN   & AUC   & LastFM/Movielens-100K/Yelp&2023\\
          & KALUR~~\cite{ma2023enhancing} &       & GNN   & Recall/NDCG & MovieLens-100K/Amazon-book/LFM-1B&2023\\
          & PCGNN~~\cite{sun2024prerequisite} &       & GNN   & HR/MRR & Udemy1/XueTangX &2024\\
          & CR-LCRP~~\cite{yu2024cr} &       & HIN   & AUC/NCDG/MRR/HR/AP & MOOCCube/MOOC review &2024\\
          & TCRKDS~~\cite{shaw2023tcrkds} &       & LSTM  & ACC/Recall/Precision/F-measure/FDR & Kaggle&2023\\
          & FRS~~\cite{zhou2018personalized}& &LSTM &AUC/Precision/RSME & ASSISTment&2018\\
          & MRCRec~~\cite{hao2023meta} &       & GCN    & HR/HR/NDCG/MRR & MOOCCube/XuetangX&2023\\
          & ConceptGCN~~\cite{alatrash2024conceptgcn} &       & GCN/Transformer    & Precision/Recall/F1 & Inspec/SemEval2017&2024\\
          & MRMLREC~~\cite{zhang2024mrmlrec} &       & GCN/GAT/LSTM    & HR/NDCG/MRR & Computer/MOOCCube&2024\\          
\cmidrule{2-7}          & RecGAN~~\cite{bharadhwaj2018recgan} & \multirow{2}[0]{*}{Unsupervised Learning} & GAN & NDCG/MRR/MAP & MPF&2018\\
          & DBNLS~~\cite{zhang2020learning} &       & DBN   & Act/Ref/Sen/Int & StarC &2020\\
\cmidrule{2-7}          & RLALS~~\cite{shawky2018reinforcement} & \multirow{10}[0]{*}{Reinforcement Learning} & AC & N/A   & N/A &2018\\
          & CSEAL~~\cite{liu2019exploiting}  & & AC & AUC/Recall/F1/MAP& Junyi & 2019\\
          & MEUR~~\cite{liang2023graph}  &       & Q-Learning & IR/HR/NDCG/MRR & MOOCCube &2023\\
          &QLearnRec~~\cite{tang2019reinforcement} &&Q-Learning&N/A&Self-Collected&2019\\
          &RILS~~\cite{raghuveer2014reinforcement} &&Q-Learning&N/A&Self-Collected&2014\\
          &RPPR~~\cite{lin2023multi} & &REINFORCE&HR/NDCG &MOOCCourse/MOOCCube &2023\\
          &HELAR~~\cite{lin2022hierarchical} & &REINFORCE&HR/NDCG&MOOCCourse/MOOCCube&2022\\
          &HRRL~~\cite{lin2022context} &&REINFORCE&HR/NDCG&MOOCCourse/MOOCCube&2022\\
          &HRL-NAIS~~\cite{zhang2019hierarchical}&&REINFORCE&HR/NDCG&MOOCCourse&2019\\
          & KRRL~~\cite{lin2024knowledge}  &       & AC &HR/NDCG&MOOCCourse/MOOCCube&2024\\
          & UPGPR~~\cite{frej2024finding}  &       & AC &NCDG/Recall/HR/Precision&COCO/XuetangX&2024\\
    \bottomrule
    \end{tabular}%
    }
  \label{tab2}%
\end{table*}%

\subsection{Knowledge Tracing}
Knowledge tracing~\cite{shen2024survey} is an educational assessment technology that tracks students' learning process and predicts their mastery of knowledge points. In a knowledge tracing scenario, students' learning process is usually recorded, including learning time, answer status, homework completion, etc~\cite{sun2024progressive,wang2024pre}.
\par According to previous studies proposed by Song et al.~\cite{song2022survey}, the problem of knowledge tracing in intelligent education systems involves three main elements: the student $S$, the exercise $E$, and the corresponding knowledge concept $C$. The interactions $X$ among these elements are the main activities in such systems. Specifically, given a student's historical exercise interactions $s \in S$, where each interaction $X_t \in X$ corresponds to an exercise $e \in E$ and denotes the correctness $a_t \in \{0,1\}$ of the result obtained at step $t$, the knowledge tracing task aims to predict the next interaction $X_{t+1}$ for a specific concept $c \in C$~\cite{song2022survey}.
\par
\begin{figure*}[htbp]
  \centering
  \includegraphics[width=0.8\textwidth]{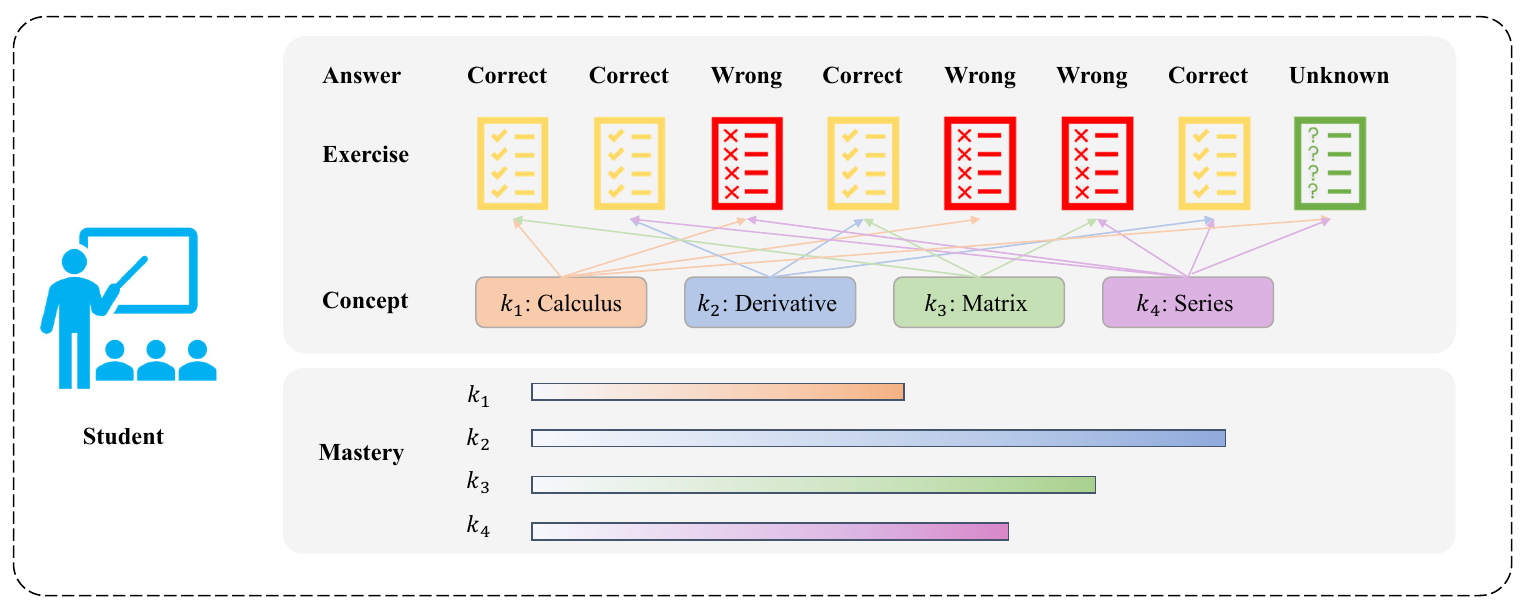}
  \caption{A simple knowledge tracing schema. Different exercises contains multiple types of concepts with different colors. Whether the exercises are correct or not will affect knowledge tracing's judgment of the student's mastery of a certain knowledge point or exercise~\cite{liu2021survey}.}
  \label{fig:KT}
\end{figure*}

 A simple schema of knowledge tracing is shown as Fig. \ref{fig:KT}. The correct rate of a student in a certain exercise on a specific knowledge point can affect the model's judgment of the proficiency level of students on this knowledge point. For example, in Fig. \ref{fig:KT}, the student's correct rate on the derivative knowledge point is 100\%, so his/her mastery level on that knowledge point is much higher than other knowledge points.

\subsubsection{Supervised Learning}
The application of supervised learning in the field of knowledge tracing is mainly focused on various kinds of neural networks, \eg LSTM~\cite{nagatani2019augmenting}, CNN~\cite{lyu2022deep}, RNN~\cite{li2023plastic}, GNN~\cite{yang2024heterogeneous} and Transformer~\cite{pu2024elakt,yang2024evolutionary}.
\par LSTM is primarily used to simulate the forgetting and learning processes of students, updating their knowledge states accordingly, thereby enabling effective knowledge tracing. A GFLDKT model based on LSTM was presented by Zhao et al.~\cite{zhao2023novel}, in which a Gating-controlled Forgetting and Learning mechanism was employed to effectively update the knowledge state and facilitate accurate prediction of subsequent student responses.
\par In addition, Nagatani et al.~\cite{nagatani2019augmenting} proposed a DFKT model considering forgetting processes of student. Specifically, DFKT applies LSTM and Neural Factorization Machine (NFM), the former is used to represent the knowledge state of the student as low dimensional dense vectors, and the latter combines student knowledge states and other related information including relevant forgetting data, to predict student performance. Compared to another LSTM-based knowledge tracing model tested on the same ASSISTment12 dataset, LFBKT~\cite{chen2022knowledge} distinguishes itself in its approach to handling forgetting behavior. Unlike the DFKT model, which integrates forgetting data as part of its input, LFBKT opts for a more nuanced treatment by incorporating a dedicated Knowledge Forgetting Layer. This strategic design choice enables LFBKT to more effectively model the dynamics of knowledge decay over time. The impact of this approach is evident in the performance enhancement, with LFBKT's accuracy improving by 6.34\% compared to the DFKT model on the ASSISTment12 dataset, demonstrating an advancement in knowledge tracing efficacy.
\par CNN also plays an essential role in knowledge tracing by being able to process the spatial sequence data. Lyu et al.~\cite{lyu2022deep} introduced a DKT-STDRL model that employs CNN to extract spatial features from students' learning sequences and LSTM to process temporal features.
\par Knowledge Graph (KG) is a crucial component of knowledge tracing, as it accounts for the interplay between a student's learning history and specific areas of expertise. By leveraging the power of the KG, models can more accurately capture learning trajectories and interactions between different points of knowledge. Yang et al.~\cite{yang2021gikt} put forth a Graph-based Interaction model for Knowledge Tracing (GIKT) constructed using Graph Convolutional Network (GCN). The proposed model addresses data sparsity and multi-skills challenges by harnessing high-order question-skill correlations, thus improving model performance.
\par Moreover, a heterogeneous graph-based algorithm called HHKST proposed by Ni et al.~\cite{ni2023hhskt}, which utilizes a GNN-based Base Feature Extractor (BFE) to extract interaction and knowledge structure features from the heterogeneous graph.
\par As a neural network architecture designed for handling sequential data, RNN also holds a significant position in the field of knowledge tracing. DKT~\cite{piech2015deep} utilized considerable amounts of artificial neurons from RNN to construct knowledge tracing model. The principle contribution of this article is the introduction of a novel method that encoding student interactions into RNN inputs and improved the AUC by 17\% to 21\% approximately compared to the Bayesian Knowledge Tracing (BKT)~\cite{corbett1994knowledge} model on three public datasets.
\par In knowledge tracing, the application of Attention mechanisms has been explored~\cite{huang2024learning,wong2024architectural}. Ghosh et al. \cite{ghosh2020context} introduced a method called Context-Aware Attentive Knowledge Tracing (AKT), which combines a flexible DNN with heuristic cognitive and psychometric models. The proposed Attention mechanism is used to dynamically adjust predictions based on contextual information about learners. This integration of Attention allows the model to adapt its predictions according to the specific context of each learner.
\par Building on the concept of attention mechanisms, Cui et al.~\cite{cui2023fine} proposed the MRTKT model as an innovative approach for knowledge tracing. Achieving an AUC of 82.23\% on the ASSISTment09 task, MRTKT slightly outperforms the AKT model, which scored 81.69\%. Unlike AKT, which emphasizes context-aware dynamic adjustments using attention mechanisms, MRT-KT employs a multi-relational attention mechanism along with a relation encoding schema to enhance its predictive accuracy.
\par Additionally, Pandey and Karypis \cite{pandey2019self} proposed the Self-Attentive Knowledge Tracing model (SAKT), which leverages a self-attentive mechanism to identify and predict the mastery level of students for specific knowledge points. The experimental results demonstrate that SAKT outperforms traditional methods and RNN-based models, achieving significantly faster performance by an order of magnitude. Furthermore, the FCIKT model~\cite{xu2024modeling} utilizes GCN to obtain higher-order relationships from the constructed heterogeneous graph, and adopts the multi-head attention mechanism to learn the rich information from individual questions.

\subsubsection{Unsupervised Learning}
Although the preceding sections of our text have listed numerous knowledge tracing methods based on supervised learning, strictly classifying scenes into specific algorithm categories (Supervised, Unsupervised, Reinforcement Learning) is often challenging. While traditional methods like BKT~\cite{corbett1994knowledge} are seen as unsupervised learning, they don't incorporate Deep Learning techniques. Hence, applying unsupervised Deep Learning in knowledge tracing is a relatively new and developing area. However, in recent years, with the rapid development of Deep Learning, some researchers have started to explore the use of unsupervised learning in knowledge tracing. For instance, Cheng et al.~\cite{cheng2022adaptkt} proposed a Autoencoder- based DKT, which combines knowledge tracing and transfer learning. The Autoencoder here is used to convert question text to high-level semantic embedding. In addition, the model also applies Bi-LSTM and Attention mechanism to capture knowledge state of student and predicts next answer of learner by Softmax.

\subsubsection{Reinforcement Learning}
Reinforcement learning is able to learn the knowledge status and level of students by rewarding or punishing through decisions and actions they make. The learning process in knowledge tracing can be considered as a sequential decision problem, students need to make different decisions according to different knowledge points, which might affect the future learning process.
\par Ding et al.~\cite{ding2020automatic} considered that some supervised learning such as LSTM or GRU, are heavily influence by NLP but not specially designed for knowledge tracing. Thus, the authors design a RL-KTNet algorithm which applies reinforcement learning to automatically generate RNN cells used in knowledge tracing. It outperforms other models employing LSTM cells in terms of AUC.
\par In addition, Long et al.~\cite{long2021tracing} proposed a model called Individual Estimation Knowledge Tracing (IEKT), which incorporates reinforcement learning for auxiliary model training. Specifically, Long et al. employed Policy Gradient and $\epsilon$-greedy to update the model parameters. Additionally, reinforcement learning is utilized to estimate student knowledge states and sensitivity towards knowledge acquisition.

\subsection{Student Behavior Detection}
In the current era, both traditional and online educational systems have been generating massive amounts of data. The extraction of useful knowledge and underlying patterns from this voluminous data can enable decision-makers to enhance teaching and learning by identifying student behaviors~\cite{prenkaj2020survey}. This educational data can be considered as an invaluable source of information that can facilitate data-driven educational research and innovation. The objective is to identify students’ behaviors like low motivation, low engagement, cheating, dropout and procrastination~\cite{el2022educational}.
\begin{figure*}[ht!]
  \centering
  \begin{minipage}{0.49\textwidth}
    \includegraphics[width=\linewidth]{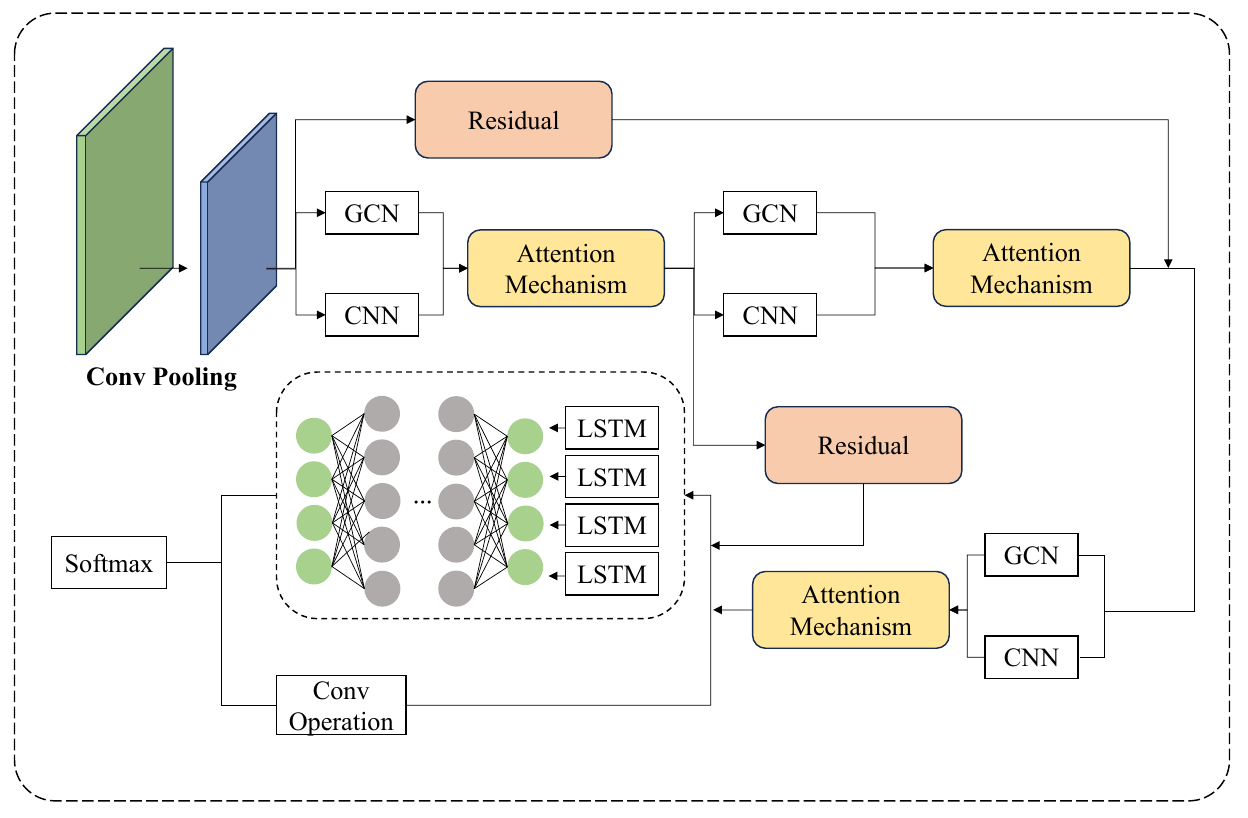}
    \caption{A simple structure of ATGCN model~\cite{qiu2023ld} for student behavior detection.}
    \label{SBD1}
  \end{minipage}\hfill
  \begin{minipage}{0.49\textwidth}
    \includegraphics[width=\linewidth]{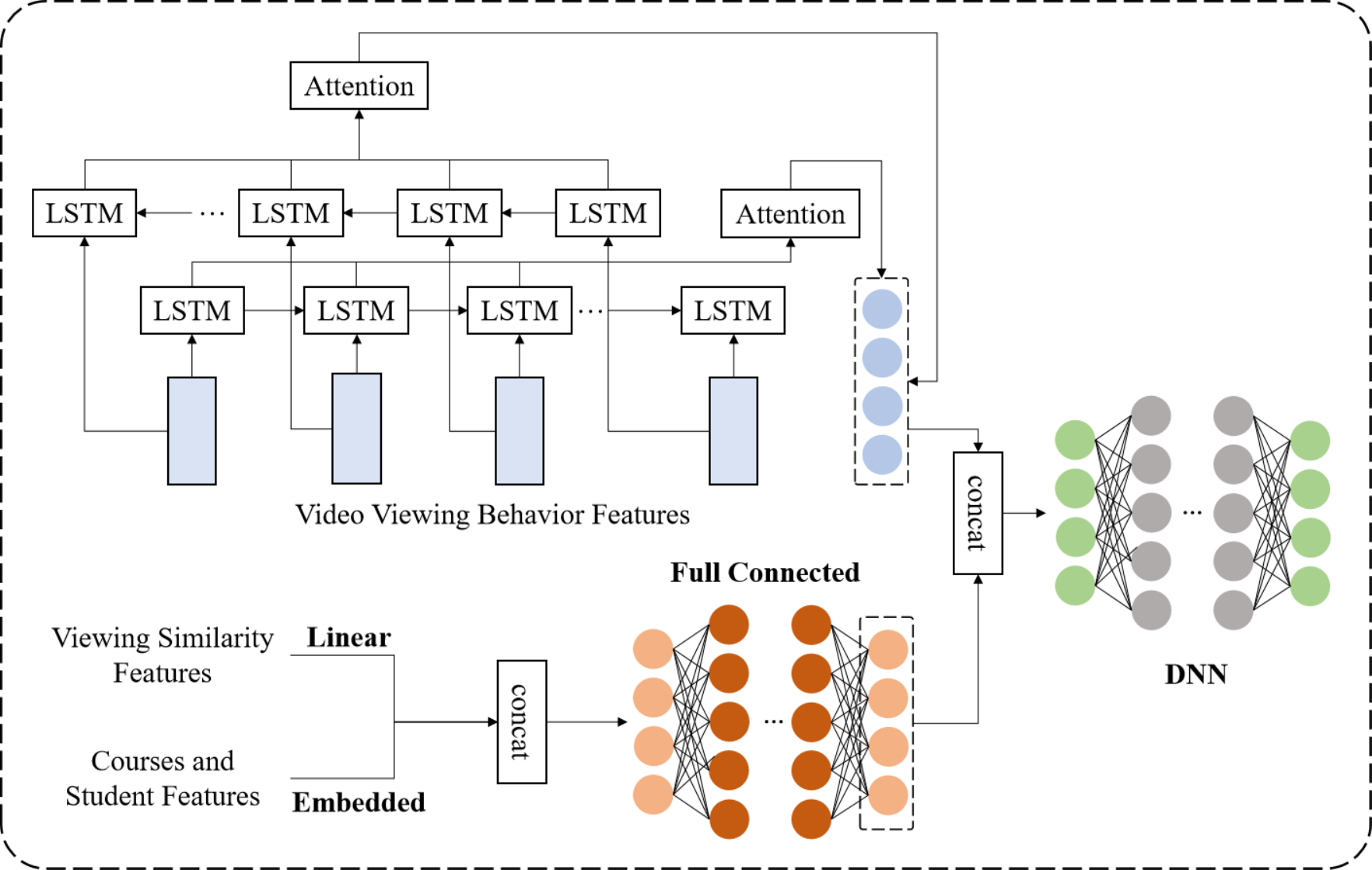} 
    \caption{A simple structure of LBDL model~\cite{liu2023learning} for dropout prediction.}
    \label{SBD2}
  \end{minipage}
\end{figure*}

\subsubsection{Supervised Learning}
The applications of supervised learning in this field are relatively diverse, variety of neural networks are applied, such as CNN and LSTM.
\par CNNs are frequently used in applications that require spatial feature extraction from image or video data due to their effectiveness in such tasks. Wang et al.~\cite{wang2023automated} implemented CNNs to analyze data obtained from motion sensors, leading to accurate recognition of 14 common classroom behaviors.
\par Leveraging CNN's superior image processing, Hdioud et al.~\cite{hdioud2023facial} developed a DCNN model for recognizing students' faces with masks during the epidemic, showcasing adaptability under challenging conditions. Concurrently, Qiu et al.~\cite{qiu2023ld} proposed an ATGCN model, also based on CNN, to identify improper behaviors like napping in classrooms. This highlights the diverse applications of CNN in educational settings, where one model focuses on facial recognition under constraints and the other on behavior detection. The ATGCN leverages both CNN and GCN as its fundamental components and employs an attention mechanism to boost its performance. A concise representation of the ATGCN structure is depicted in Fig. \ref{SBD1}.
\par In education scenarios, student behaviors extend beyond simple actions such as napping and inattentiveness. An important part of this field is the detection of dropout tendencies, integral to understanding the full scope of student behaviors. Feng et al. \cite{feng2019understanding} proposed a Context-aware Feature Interaction Network (CFIN) model for predicting students' dropout behaviors in MOOCs. The CFIN model incorporates a context smoothing technique to enhance the feature values across different contexts and employs an attention mechanism to integrate user and course information within the modeling framework.
\par Furthermore, Kumar et al.~\cite{kumar2023ensemble} developed an EDLN model that combines CNN and an attention mechanism for student dropout detection in online courses. The EDLN model employs ResNet-50 to extract local high-dimensional features and uses Faster R-CNN to analyze hidden long-term memory features in time series data. This model demonstrates notable performance, achieving an accuracy of 97.5\% with a $5\times7$ time series matrix as input.
\par Other deep neural networks can be also employed for dropout prediction. The LBDL model, as proposed by Liu et al.~\cite{liu2023learning}, integrates Bi-LSTM and a multi-head attention mechanism to analyze time series information extracted from video-based study behaviors. This model demonstrates performance with AUC and F1 scores of 82.39\% and 74.89\%, respectively, on the MOOCCube dataset. A simple structure of LBDL is shown as Fig. \ref{SBD2}. Subsequently, the authors demonstrate the exceptional performance of this model.
\par With the advancement of online education, the detection of cheating has become a critical area of focus for EDM researchers. Alsabhan et al.~\cite{alsabhan2023student} developed the CDLSTM model, which is based on the LSTM. This model incorporates a dropout layer, a dense layer, and the Adam optimizer. Owing to the LSTM's ability to effectively process sequential data, the CDLSTM model achieved a notable accuracy of 90\% in identifying cheating behaviors, using students' online test activity logs as input.

\subsubsection{Unsupervised Learning}
In this section, we delve into the application of unsupervised learning within EDM, particularly in detecting student behaviors. This approach, a facet of Deep Learning, processes educational data to autonomously identify patterns in student engagement, providing insights without relying on predefined labels or categories.
\par Stenton et al.~\cite{stenton2021fine} introduced a FTGAN which means Fine-Tuning GAN to predict the attrition rate of the student. The authors demonstrate that the more epochs the GAN classifier model is trained, the more its accuracy shows a certain level of increase.
\subsubsection{Reinforcement Learning}
While exploring the application of various algorithms in detecting student behaviors, it's crucial to consider the complexity and multifaceted nature of these behaviors. Reinforcement learning, though powerful, faces challenges in this domain:
\begin{itemize}
    \item The complexity of student behavior, influenced by diverse factors like social background and personal circumstances, may not align well with the straightforward state-action framework typically used in reinforcement learning.
    \item Defining appropriate rewards and punishments in the context of student behaviors and their long-term outcomes, such as dropout rates, presents a significant challenge. The dichotomy between short-term and long-term consequences complicates the application of reinforcement learning.
\end{itemize}
Based on the outlined reasons, our study does not include literature on the application of reinforcement learning for student behavior detection. However, we acknowledge the potential value and relevance of attempts to explore this method in such contexts.

\subsection{Performance Prediction}
Performance prediction~\cite{rastrollo2020analyzing} refers to the use of various data and analytic techniques to predict student performance in certain tasks or areas, such as test scores, academic performance, course completion rates, etc. The difference between it and the knowledge tracking task is that performance prediction focuses on predicting students' overall future tasks or test performance based on historical learning data, while knowledge tracing focuses on students' understanding and mastery of specific knowledge concepts during the learning process.

\par By predicting student performance, teachers and educational institutions can better understand students' learning and needs to provide more effective support and guidance. Predictions can also help students understand their performance and potential difficulties and take steps to improve learning outcomes.
\par Furthermore, cognitive diagnosis~\cite{wang2020neural} is a related concept wherein the aim is to identify students' level of mastery in specific knowledge domains by analyzing their performance on exercise records. This analysis facilitates providing tailored guidance for their subsequent studies~\cite{gao2021rcd, tong2022incremental}. In contrast, performance prediction focuses more on predicting students' overall scores on tests. The former analyzes knowledge dimensions, while the latter emphasizes general ability. Cognitive diagnosis outputs concept proficiency levels, whereas performance prediction directly predicts total scores.

\subsubsection{Supervised Learning}
Supervised learning algorithms can provide teachers and students with useful information to help them better understand student performance and needs. In addition to classic algorithms like SVM, Deep Learning models such as MLP, CNN, RNN, and LSTM have been applied to predict student performance in various educational contexts.
\par Nayani et al.~\cite{nayani2023combination} introduced a hybrid model called CRN which is a combination of CNN and RNN to predict the students' grade, and improves the performance by tuning hyperparameters through Galactic Rider Swarm Optimization (GRSO) algorithms.
\par In addition, Neha et al.~\cite{neha2023deep} presented a SDPNN model that applied a linear classifier-based DNN to predict student academic performance. There are two hidden layers with 300 neurons defined. The activation function is ReLU and Softmax.
\begin{equation}
    \text{softmax}(z_i)=\frac{e^{z_i}}{\sum_{j=1}^{K} e^{z_j}}, \quad i=1,2,...,K,
    \label{eq:9}
\end{equation}
where $z_i$ represents the $i_{th}$ element of the input vector, and $K$ is the number of categories.
\par Kukkar et al.~\cite{kukkar2023prediction} proposed a SAPP system that utilize four layers of stacked LSTM, Random Forest and Gradient Boosting. Here, LSTM is used to extract features while Random Forest and Gradient Boosting are applied to predict. The proposed system has an accuracy rate of 96\% and performs well to some extent.
\par Furthermore, Chen et al.~\cite{chen2023prediction} introduced the MSH-MD model, which also utilizes LSTM. The core of MSH-MD is its self-attention mechanism, integrating LSTM's capacity for processing time series with the efficient feature extraction of the self-attention mechanism. Compared to the SAPP model \cite{kukkar2023prediction}, the MSH-MD focuses more on predicting student performance through pattern differences, whereas the SAPP model emphasizes feature extraction more prominently.

\subsubsection{Unsupervised Learning}
\par Although the grade prediction usually relies on labeled datasets for supervised learning, some unsupervised learning algorithms, such as GAN, can also be used for grade prediction tasks. GAN can perform grade prediction by taking a student's historical grades as input and using generators and discriminators to generate predicted values for future grades. The generator can use the student's prior grades and other relevant factors to generate predictions of future grades, and the discriminator is used to determine whether the generated predictions are similar to the true grades.
\par Sarwat et al.~\cite{sarwat2022predicting} proposed a model combined with Conditional GAN (CGAN) and Deep-Layer-based SVM to predict students’ grades according to school or home tutoring. CGAN was used to generate performance score data to address the issue of small dataset size, and the model using a combination of CGAN and SVM was experimentally shown to have a positive effect on the prediction results.
\subsubsection{Reinforcement Learning}
Generally speaking, while reinforcement learning may not be traditionally used for student performance prediction, it can be employed to optimize a student's learning path by setting reward strategies based on student behavior. In this context, reinforcement learning would also involve assessing the current performance of students and considering potential improvements.
\par Dorça et al.~\cite{dorcca2013comparing} proposed an ADSLS model to automatically detect and precisely adjust students' learning styles. An important part of this is that the model predicts and assesses student performance on a point and rewards performance while updating learning strategies. The proposed method's efficacy and efficiency have been demonstrated by the results.
\par Zhuang et al. ~\cite{zhuang2022fully} introduced an NCAT model that utilizes reinforcement learning to enhance the effectiveness of e-testing systems. The NCAT model employs deep reinforcement learning to dynamically optimize test terms based on given conditions. With the proposed algorithms, the e-testing system can provide more comprehensive and accurate performance predictions.

\subsection{Personalized Recommendation}
In the current era of explosive growth in online information, recommendation systems~\cite{khanal2020systematic,ain2024learner} undoubtedly offer an effective means of addressing this issue and providing necessary assistance to individual users. Even outside the educational context, recommendation systems remain one of the most widely studied technological approaches. Zhang et al.~\cite{zhang2019deep} gave a detailed definition of a recommendation system that: Assuming the presence of $M$ users and $N$ items, we denote the interaction matrix and predicted interaction matrix as $R$ and $\hat{R}$, respectively. The user preference for item $i$ is represented as $r_{ui}$, while the predicted score is denoted by $\hat{r}_{ui}$. There will also be two partially observed vectors, one represents a specific user $u$, \ie $r^{(u)}=\{ r^{u1}, ..., r^{uN}\}$. The other one represents a specific item $i$, \ie $r^{(i)}=\{ r^{1i}, ..., r^{Mi}\}$.
\par
\begin{figure*}[htbp]
  \centering
  \includegraphics[width=0.8\textwidth]{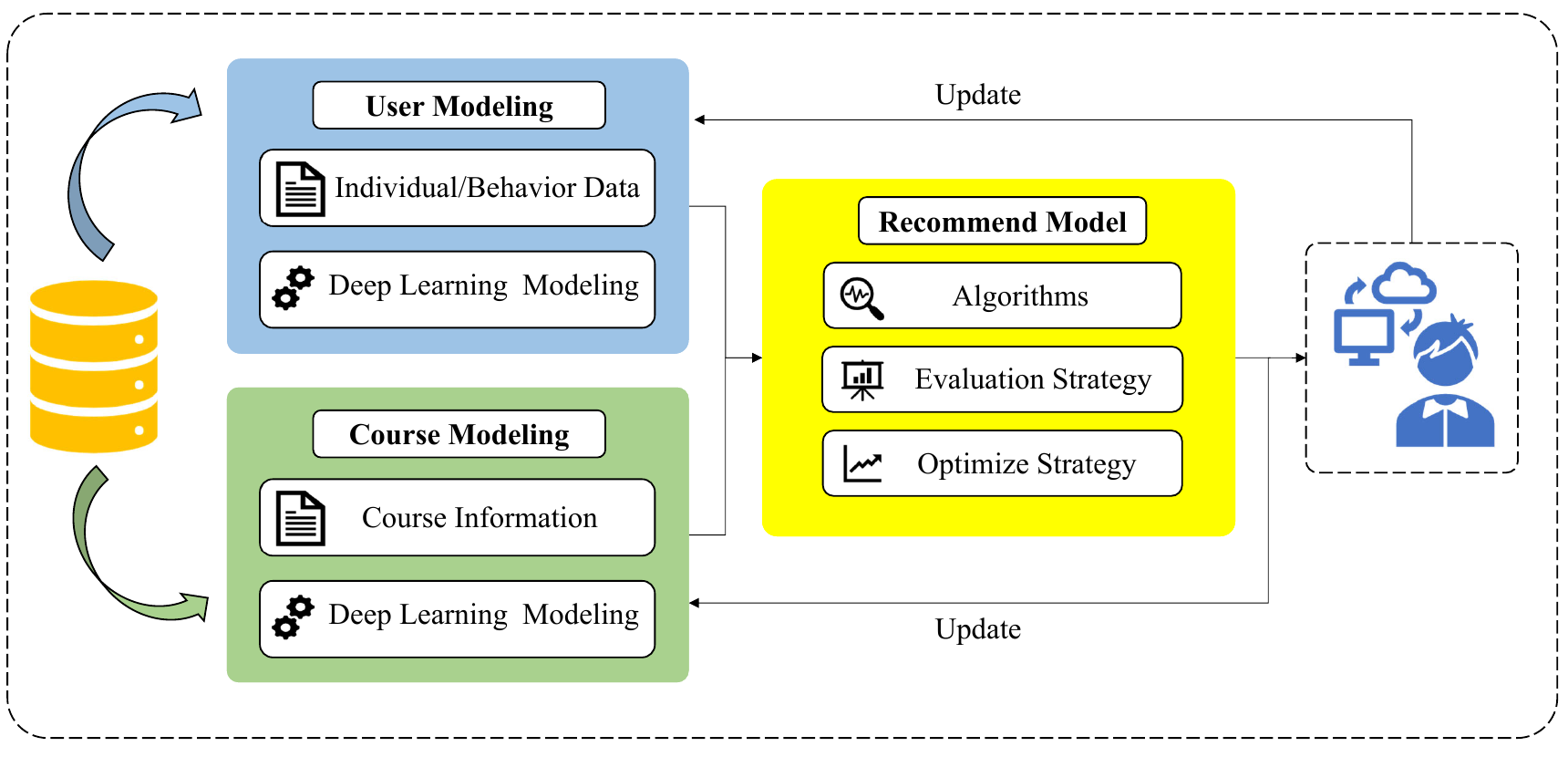}
  \caption{A general process framework for personalized recommendation in EDM. The recommendation system uses user data and course data for modeling. After recommending to users, the associated data between users and courses will be updated to improve accuracy.}
  \label{fig:PS}
\end{figure*}

 In the education scenarios, the item is replaced with educational resources such as courses. The primary objective of a course recommendation system is to suggest the most suitable course to a user at time $t + 1$, taking into account their past learning activities and learner profiles prior to time $t$. The primary challenge faced by such recommendation systems is to provide personalized recommendations by precisely depicting and conceptualizing user inclinations through analysis of user data~\cite{liu2022review}. Fig. \ref{fig:PS} demonstrates a simple schema of personalized recommendation.
\subsubsection{Supervised Learning}
The application of supervised learning in course recommendation~\cite{sakboonyarat2022applied} focuses on a wide variety of neural network models, due to the greater flexibility and expressiveness of neural networks, which can better handle user behavior sequences and nonlinear features. At the same time, neural networks can automatically extract features from the data to more accurately predict users' interests in the recommendation process.
\par Some optimized neural network models play a role in the field of personalized recommendation. SODNN, a novel model consisting of synchronous sequences, heterogeneous features and DNN has been introduced by Safarov et al.~\cite{safarov2023deep}. At the same time, to solve the cold-start problem, \ie Large errors caused by missing user data during the initialization of the recommendation system, the authors tried to concatenate additional features to overcome it.
\par Furthermore, classical neural networks can also be employed in recommendation systems. In contrast to the conventional positive sequence modeling approach, Gao et al.~\cite{gao2022online} proposed a novel CSEM-BCR model that adopts negative sequence modeling. Specifically, this model constructs the course-learning sequence as a negative sequence pattern, which the negative term refers to the principle that students should not select or operate courses inappropriately. The negative sequence pattern is then processed using CNN for feature learning, which generates a list of recommended courses for each user. The suggested approach presents a fresh perspective on personalized recommendation and offers a potential solution to the issue of recommending courses to learners with different needs and preferences.
\par KG is an important part of personalized recommendations. It can effectively address the sparsity issue in recommender systems. An ARGE model based on multiple paths RNN encoder is proposed by Zhao et al.~\cite{zhao2023agre}. The model solves the problem that traditional RNNs do not consider the association between paths for encoding, and the AUC and Precision in the experimental outcomes demonstrate the model's capacity to effectively address the issue of sparse interaction between users and items.
\par GNN, as a neural network model specifically for processing graph structures, has also been applied by many scholars to deal with recommender systems~\cite{sun2024prerequisite,yu2024cr}. Ma et al.~\cite{ma2023enhancing} presented a KAUR model that applies GNN to learn node representations for each node in the collaborative KG. The model treats the node information and its neighboring node information that has been propagated as positive contrastive pairs, and then leverages contrastive learning to improve the quality of the node representations.
\par Zhou et al. \cite{zhou2018personalized} proposed a Full-path Recommendation System (FRS) based on LSTM and a clustering algorithm. The clustering algorithm is utilized to categorize learners based on their similar learning features, which in turn helps classify the learning paths according to the previous results. Consequently, this approach effectively addresses the cold start problem. LSTM is employed for predicting learning performance, and if the result is unsatisfactory, the system will select the most relevant learning path for users based on their individual learning features.
\par Some researcheres leveraged the hybrid methods~\cite{zhang2024mrmlrec} to simultaneously address multiple issues or challenges in course recommender systems. For example, Alatrash et al.~\cite{alatrash2024conceptgcn} proposed a ConceptGCN framework to enhance the e-learning experience. It combines propagation-based GCN with KG to effectively represent items (\eg courses, videos, and knowledge concepts), and then adopts Transformer to better capture the importance of relations between different knowledge concepts. Finally, it uses these enhanced representations to construct learner models, facilitating more personalized recommendations of associated knowledge concepts.

\subsubsection{Unsupervised Learning}
In recommender systems, user preferences and behaviors are often incomplete and inaccurate, and tagging data is very difficult and expensive to collect. On this basis, unsupervised learning is a great means to achieve personalized recommendations. Unsupervised learning can extract potential interests and preferences from users' historical behaviors and infer the similarity and interest relevance of users through clustering, and feature learning to achieve personalized recommendations.
\par Bharadhwaj et al.~\cite{bharadhwaj2018recgan} introduced a hybrid RecGAN model based on GAN and RNN. The generator and discriminator are both constructed with GRU-based RNN. The generator is then allowed to play a Mini-Max game with the discriminator, \ie there exists a true distribution $D_{real|t}$ in time index $t$, and a probability distribution generated by the generator $D_{gen|t}$. The goal of this minimal-maximization game is to minimize the generation error of the generator, while maximizing the discriminator's ability to distinguish false ratings from true ratings.
\par Unsupervised learning can also be applied to classify different types of learning styles to recommend the most suitable education resources to distinct groups. A DBN-based DBNLS model proposed by Zhang et al.~\cite{zhang2020learning} is used to detect and classify learning styles. The core component of DBNLS is the multilayer RBM and Back Propagation network layer, where Back Propagation is used to fit the DBNLS model and to fine-tune and train the DBN.

\subsubsection{Reinforcement Learning}
The application of reinforcement learning in personalized recommendation has high research value and development prospects. Reinforcement learning algorithms can learn the optimal recommendation strategy based on learner' feedback to improve the recommendation performance.
\par A cognitive structure enhanced framework for adaptive learning named CSEAL was proposed by Liu et al.~\cite{liu2019exploiting}, to achieve personalized learning path recommendation. This framework views the learning path as a Markov Decision Process (MDP) and applies actor-critic to identify appropriate learning projects for individual learners. CSEAL comprehensively considers the learner's knowledge level and the knowledge structure of the learning project. Experimental results demonstrate that CSEAL can enhance learners' efficiency compared with current adaptive learning methods.
\par Liang et al.~\cite{liang2023graph} presented a MEUR model based on Graph Convolution Network and reinforcement learning. The author considered the learning process as a MDP. Applying a user-centric reasoning method and maximizing cumulative reward $G$ by actor-critic algorithm, $G$ is defined as below:
\begin{equation}
    G(\theta)=\frac{1}{m} \sum_{u=1}^m \sum_{a,t=0} \pi (a|s_t,A(s_t))\gamma^tR(s_{t+1}),
    \label{eq:10}
\end{equation}
where $\theta$ is the parameter of the Actor network, $m$ is the number of samples, $u$ is the sample index, $a$ and $t$ are the time step and action indexes, $\pi(a|s_t,A(s_t))$ is the probability of selecting action $a$ in state $s_t$, $R(s_{t+1})$ is the immediate reward obtained in state $s_{t+1}$, and $\gamma$ is the tuning parameter.
\par It is important to propose explainable recommendation methods in MOOC for interpreting the recommendation results. To this end, Lin et al.~\cite{lin2024knowledge} developed a KRRL model, combined KG and self-supervised reinforcement learning, for explainable MOOC recommendation. Specifically, a multi-granularity representation learning method enriches the perceptual information of semantic interactions in the KG, and a self-supervised reinforcement learning method guides the path reasoning over the KG, and finally recommends appropriate courses to the target learner. The experimental results show that KRRL performs better than some comparison methods, in terms of the recommendation accuracy and explainability. Similarly, the UPGPR model~\cite{frej2024finding} also utilizes the KG and AC algorithm to implement explainable recommendation in MOOC.
\par Besides, reinforcement learning algorithms can be used to model learner preferences. For example, Zhang et al.~\cite{zhang2019hierarchical} are the first to adopt the Hierarchical Reinforcement Learning (HRL) to modify learner profiles. Furthermore, the HRRL algorithm~\cite{lin2022context} integrates Recurrent Reinforcement Learning (RRL) with HRL as a profile reviser, which iteratively revises learner profiles to assist the MOOC recommendation model. Lin et al.~\cite{lin2023multi} also proposed a RPPR model, in which a multi-scale reinforcement learning method constructs multi-dimensional learner profiles, thereby dealing with the adverse effects of insufficient semantic information on learner modeling.

\section{Datasets and Processing Tools}
\subsection{Datasets}
\begin{table*}[htbp]
  \centering
  \caption{Dataset Information}
      \label{tab3}
   \begin{threeparttable}
   \scriptsize
    \begin{tabularx}{\textwidth}{>{\raggedright\arraybackslash}p{2.5cm}>{\raggedright\arraybackslash}p{6cm}>{\raggedright\arraybackslash}X>{\raggedright\arraybackslash}p{3cm}}
    \toprule
    \textbf{Dataset Name} &\textbf{ Description} & \textbf{Application Scenario} & \textbf{Literature Applied} \\
    \toprule
    ASSITments\tnote{1} & A knowledge tracing dataset based on knowledge decomposition theory in education, including problem-solving records. Aimed for developing and evaluating intelligent education systems. & Used for researching and evaluating intelligent education systems based on knowledge decomposition theory, such as knowledge tracing and personalized recommendation, using machine learning and Deep Learning techniques. &~\cite{piech2015deep}~\cite{zhao2023novel}~\cite{ni2023hhskt}~\cite{lyu2023dkt}~\cite{li2023plastic}\par~\cite{zhang2022online}~\cite{chen2022knowledge}~\cite{xiao2023knowledge}~\cite{lyu2022deep}~\cite{cui2023fine}\par~\cite{zhuang2022fully}~\cite{yang2021gikt}~\cite{wang2020neural}~\cite{long2021tracing}\par~\cite{ghosh2020context}~\cite{pandey2019self}~\cite{han2023deep}\\
    Junyi Academy Online Learning Activity Dataset\tnote{2} & Junyi dataset is an educational dataset that contains a large amount of online learning behavior data of primary and secondary school students, including learning and answering records, learning behavior features, and other information. & The dataset is designed to support data mining task, \eg knowledge tracing. &~\cite{zhao2023novel}~\cite{ni2023hhskt}~\cite{lyu2023dkt}~\cite{li2023plastic}~\cite{liu2019exploiting}\par~\cite{zhou2018personalized}~\cite{long2021tracing}\\
    KDD Cup 2010\tnote{3} & This dataset contains logs of student interactions with computer-aided-tutoring systems, including problem-solving transactions. Key terms include problem, step and knowledge component. & It can be applied to the development and evaluation of intelligent educational systems, personalized recommendation, knowledge tracing and Deep Learning-based educational technologies. &~\cite{zhang2022online}~\cite{montero2018does}~\cite{wilson2016estimating}~\cite{wilson2016back}\par~\cite{fu2021clsa}\\
    MOOCCube\tnote{4} & MOOCCube is an open data warehouse for NLP, KG, and data mining researchers in large-scale online education. It includes 706 real online courses, 38,181 instructional videos, 114,563 concepts, and 199,199 MOOC users' hundreds of thousands of course selection and video-watching records. & The data set can be used to study learner behavior patterns, which can be utilized for personalized recommendation.  &~\cite{zhang2023kgan}~\cite{hao2023meta}~\cite{wang2022multi}~\cite{liang2023graph}~\cite{lin2021adaptive}\par~\cite{lin2022hierarchical}~\cite{lin2022context}~\cite{liu2023learning}\\
    xAPI-Educational Mining Dataset\tnote{5} & This dataset contains 480 student records and 16 features, including demographic attributes (gender, nationality), academic background information (educational stage, grade level, section), and behavioral indicators (raising hand in class, accessing resources, parental survey responses, school satisfaction). & xAPI-educational mining dataset can be used to develop and evaluate learning analytics models, to implement specific tasks such as personalized recommendation, performance prediction and student behavior detection in the education domain. &~\cite{bharadhwaj2018recgan}~\cite{uzel2018prediction}~\cite{buraimoh2021predicting}\\
    Open University Learning Analytics Dataset\tnote{6}  & The Open University Learning Analytics Dataset is a comprehensive educational dataset that includes over 300,000 students' demographic information, course information, interaction data, and achievement data from seven courses, and has been used in various educational research studies. & Predicting student success and dropout rates in higher education, as well as understanding learning behavior patterns in online courses. So as to develop personalized recommendation and performance prediction system. &~\cite{kukkar2023prediction}~\cite{shaw2023tcrkds}~\cite{arashpour2023predicting}\\
    Canvas Network Dataset\tnote{7} & Online courses data collected from Canvas Network platform, including course metadata, enrollment, and interaction logs. & Predicting and understanding student performance in online courses, and designing personalized recommendation and student behavior detection model. &~\cite{xing2019dropout}~\cite{zhang2022novel}~\cite{sahebi2018student}~\cite{li2019personalized}\par~\cite{assami2022implementation}\\
    Learn Moodle\tnote{8}  & The Learn Moodle dataset is a collection of student activity logs, discussions, quizzes, and more from various Moodle courses. It is used for researching student behavior patterns, learning outcomes, course design, and Moodle platform functionality in online education. & This dataset is mainly used to study student behavior patterns, learning outcomes, course design, and Moodle platform functionalities in online education, aiming to improve teaching quality and efficiency in online education. &~\cite{romero2013meta}\\
    XuetangX\tnote{9} & XuetangX is a Chinese dataset, which covers courses and learning behavior data of Chinese users on the online platform of XuetangX. This includes student registration information, course access records, video viewing behavior, homework submission, discussion participation, etc.& The data set can be used in the research and application of learner behavior analysis, knowledge tracing and personalized recommendation. &~\cite{zhang2019hierarchical}~\cite{hao2023meta}~\cite{liang2023graph}~\cite{lin2021adaptive}~\cite{lin2022hierarchical}\par~\cite{lin2022context}~\cite{wang2017deep}~\cite{feng2019understanding}\\
    EdNet\tnote{10}  & EdNet is a large-scale educational dataset collected from Santa, an AI tutoring service with over 780K users. It contains multivariate data on student interactions across platforms, including materials consumed, responses given, and time spent on learning activities. The key properties are its large scale, detailed student behavior data, and collection from a deployed system with numerous real users. & EdNet enables personalized recommendation, knowledge tracing and educational optimization through its large-scale, detailed data on student behaviors and interactions.&~\cite{cui2023fine}~\cite{yang2021gikt}~\cite{long2021tracing}~\cite{pu2024elakt}~\cite{huang2024learning}\par~\cite{yang2024evolutionary}\\
    \bottomrule
    \end{tabularx}

      \begin{tablenotes}
        \footnotesize
        \item[1] https://sites.google.com/view/assistmentsdatamining/home
        \item[2] https://www.kaggle.com/datasets/junyiacademy/learning-activity-public-dataset-by-junyi-academy
        \item[3] https://pslcdatashop.web.cmu.edu/KDDCup
        \item[4] http://moocdata.cn/data/MOOCCube
        \item[5] https://www.kaggle.com/datasets/aljarah/xAPI-Edu-Data
        \item[6] https://analyse.kmi.open.ac.uk/open\_dataset
        \item[7] https://doi.org/10.7910/DVN/GVLFXO
        \item[8] https://research.moodle.org
        \item[9] http://moocdata.cn/data/user-activity
        \item[10] https://github.com/riiid/ednet
      \end{tablenotes}
    \end{threeparttable}
\end{table*}


To provide a more comprehensive overview of commonly used public datasets in educational settings, we have curated a selection of datasets (\eg ~\cite{romero2013meta, mihaescu2021review}), as presented in Table \ref{tab3}. These datasets have been widely employed in various educational research studies and have contributed significantly to the advancement of the field. Table \ref{tab3} offers essential information about each dataset, including its name, URL, description, application scenarios, and literature applied.

\par We classify the collected datasets by source into three categories: Datasets Used for Competitions, Datasets from Online Education Platforms, and Open Data Repository.
\subsubsection{Datasets Used for Competitions}
\textbf{ASSISTments} The ASSISTments dataset has many versions, such as Assistment 2009, 2012 and 2017, which have been used in several EDM competitions to promote research and development on educational technology and learning analytics. One of the most well-known competitions is ASSISTments Data Mining Competition.

\textbf{KDD Cup 2010} This dataset is used in the KDD Cup 2010 EDM Challenge, which requires participants to use the student interaction logs contained in the provided dataset to train a new learning model and ultimately judge the results based on how accurately their model predicts student responses to new questions. This dataset is therefore also widely used for problems such as Knowledge Tracking.

\subsubsection{Datasets from Online Education Platforms}
\textbf{Junyi Academy Online Learning Activity Dataset} The dataset was sourced from the Junyi Academy online learning platform, which provides personalized learning resources and support for students. The dataset consists of over 72,000 students' records of more than 16 million practice attempts on the platform within one year, from August 2018 to July 2019.

\textbf{Canvas Network Dataset} The Canvas Network dataset is derived from the Canvas Network platform, which provides educational institutions and educators with the tools and resources to create and deliver online courses. The dataset contains a lot of course information as well as user interaction logs, etc.

\textbf{Learn Moodle} It is derived from the Moodle Learning Management System, a widely used open source online learning platform that supports educational institutions and teachers in creating, managing and delivering online courses.

\textbf{XuetangX} The XuetangX dataset is derived from the online education platform XuetangX. It is a well-known online education platform in China, established in 2013, which provides MOOCs and other online learning resources, video viewing behavior, assignment submission and discussion participation, etc.

\textbf{EdNet} It is an extensive educational dataset amassed by Santa, an AI-based online teaching platform with a user base of 780,000 in South Korea. It encompasses two years' worth of student learning behavior data and offers a plethora of information on student-system interaction, including knowledge tracking, cognitive processes, learning analysis data, and comprehensive records of students' online learning activities.

\begin{table*}[htbp]
  \centering
  \caption{Dataset Processing Tools Information}
  \label{tab4}
  \begin{threeparttable}
    \begin{tabularx}{\textwidth}{>{\raggedright\arraybackslash}p{3cm}>{\raggedright\arraybackslash}X}
    \toprule
    \textbf{Tool Name} &\textbf{ Description}\\
    \toprule
    GISMO\tnote{1} & Graphical Interactive Student Monitoring Tool for Moodle (GISMO) provides an intuitive graphical interface to visually display information about student learning activities, engagement, grades, and progress. \\
    Meerkat-ED\tnote{2} & Meerkat-ED is an educational data analysis tool designed to help educators and researchers conduct in-depth analysis of student learning data. It generates comprehensive summaries of participants' engagement in discussion forums, illustrating their interactions, identifying discussion leaders and peripheral students, and providing various additional insights\\
    Datashop\tnote{3}  &DataShop is a collection of datasets and tools for educational mining. It not only collects and provides a large amount of educational data, including students' interaction data, response records, learning trajectories, assessment results, etc. in online learning environments, but also provides a series of tools and APIs for processing and analyzing these data, facilitating researchers to perform data mining, model building and evaluation.\\
    SNAPP\tnote{4} &Social Networks Adapting Pedagogical Practice (SNAPP) is a software tool that allows users to visualize the network of interactions generated by posts and responses in discussion forums. The network visualization of forum interactions will provide teachers with the opportunity to quickly identify patterns of user behavior.\\
    LOCO-Analyst\tnote{5} &LOCO-Analyst is an advanced educational tool designed to support teachers in evaluating and improving web-based learning environments. It provides valuable insights and feedback covering all aspects of the learning process, helping educators enhance the content and structure of their online courses and providing targeted feedback and recommendations so they can optimize course design and pedagogy.\\
    StREAM\tnote{6} &StREAM is a Student Engagement Analytics Platform which is also a predictive algorithm developed by Solutionpath that provides educators with visualization of student engagement levels and identification of students who need help with certain tasks. For students, it provides information about the progress and status of their learning, enabling students and educators to adjust their learning or teaching strategies in a timely manner to improve teaching effectiveness.\\
    \bottomrule
    \end{tabularx}
      \begin{tablenotes}
        \footnotesize
        \item[1] https://gismo.sourceforge.net
        \item[2] http://www.reirab.com/MeerkatED
        \item[3] https://pslcdatashop.web.cmu.edu/index.jsp
        \item[4] https://web.archive.org/web/20120321212021
        \item[5] http://jelenajovanovic.net/LOCO-Analyst
        \item[6] https://www.solutionpath.co.uk/stream
      \end{tablenotes}
    \end{threeparttable}
\end{table*}

\subsubsection{Open Data Repository}
\textbf{MOOCCube} It is an open data repository for researchers in natural language processing, KG, and data mining related to large-scale online education, containing 706 real online courses, 38,181 instructional videos, 114,563 concepts, hundreds of thousands of course selections from 199,199 MOOC users, video viewing records, and a supplemental repository containing hundreds of thousands of academic paper resources related to in-class concepts. The concept description data is from Baidu and Wikipedia, and the course data and student behavior data are from XuetangX. Academic paper data is obtained from Aminer, a large-scale academic search engine.

\textbf{xAPI-Educational Mining Dataset} This Dataset refers to a dataset based on the Experience API (xAPI) standard. xAPI is an open learning technology specification for recording learner behavior and interaction data in a variety of learning environments. The links in the table point to a publicly available xAPI-compliant database, Students' Academic Performance Dataset, stored on Kaggle, which contains 480 samples with sixteen features.

\subsection{Processing Tools}
There are many commonly used tools for processing and analyzing educational data in educational scenarios to provide powerful support and insight for educational researchers and teachers. This section will focus on several representative tools, including LOCO-Analyst, Datashop, SNAPP, GISMO, and Meerkat-ED etc. These tools have unique functions and features in educational data processing that can help education practitioners better understand the learning process and optimize instructional design and practice. By using a combination of these tools, educational researchers and teachers can deeply analyze data on student learning behaviors, engagement, and learning outcomes, and gain valuable insights from them. The names, links, and detailed descriptions of these tools are provided in Table \ref{tab4}. 


\section{Practical Challenges}
While EDM holds significant promise, its practical application is fraught with challenges. There are five practical challenges in EDM and strategies to address them.

\subsection{Data Quality and Integration}
Educational data is often heterogeneous, coming from various sources such as Learning Management Systems (LMS), online learning platforms, and classroom interactions. These sources vary in format, structure, and quality, making integration a complex task. Additionally, missing, incomplete, and inaccurate data can further complicate the analysis.
Educational institutions should implement robust data integration platforms that facilitate the aggregation of data from multiple systems. The use of Extract, Transform, Load (ETL) processes can help in standardizing data formats and ensuring that disparate datasets are harmonized. Institutions should also establish data governance frameworks that include regular data audits and validation processes to maintain high data quality. Employing data imputation techniques, such as multiple imputation~\cite{rubin2018multiple} to handle missing data, can effectively manage gaps in data. Besides, Generative models~\cite{harshvardhan2020comprehensive} (\eg GANs) could be useful for creating synthetic educational data that maintain the statistical properties of real-world data while ensuring student privacy.

\subsection{Evaluation and Validation}
Ensuring that models and insights derived from EDM are accurate, reliable, and generalizable across different educational contexts is a significant challenge. The diversity of educational settings, curricula, and student demographics means that findings from one context may not necessarily apply to others. Additionally, maintaining the effectiveness of EDM tools over time requires continuous evaluation and refinement, which can be resource-intensive.

It is crucial to conduct extensive pilot studies and validation processes across multiple contexts before fully implementing EDM models. This can involve testing models in different types of schools, varying geographic locations, and among diverse student populations to ensure that the findings are robust and applicable in a wide range of scenarios. Transfer learning techniques~\cite{niu2020decade} are useful in adapting models to new contexts. They can improve model performance in different educational settings without the need for extensive retraining.

Moreover, institutions should also develop metrics and benchmarks for evaluating the success of EDM initiatives. These could include student performance indicators, engagement levels, and teacher satisfaction. Regularly reviewing these metrics can help in adjusting strategies and ensuring that EDM tools continue to meet the needs of educators and students. For example, it is necessary to define and standardize evaluation metrics that effectively reflect the performance of Deep Learning models in EDM. To this end, deep meta-learning~\cite{huisman2021survey} could be employed to assess and compare the performance of different algorithms on EDM tasks.

\subsection{Scalability and Real-time Processing}
Educational data can be vast, especially in large institutions or online learning platforms, making it difficult to develop algorithms that scale efficiently. To tackle scalability issues, institutions can leverage scalable data processing frameworks such as Apache Hadoop and Spark. These frameworks can manage large datasets efficiently and are well-suited for the distributed nature of educational data. Cloud computing services offer flexibility and scalability, reducing the need for extensive on-premises hardware.

Real-time processing is imperative for adaptive learning systems, which must respond dynamically to student inputs. However, this necessitates substantial computational resources and sophisticated algorithms. Employing stream processing frameworks such as Apache Kafka enables the real-time analysis of data as it is generated, thereby facilitating the development of adaptive learning environments. It is essential to invest in robust technological infrastructure and continually update it to keep pace with advancements.

\subsection{Interpretability of Models}
Deep learning techniques often yield models that are complex and difficult to interpret. For educators and administrators to trust and effectively use these models, they need to understand how decisions are made. Additionally, the insights generated by EDM must be actionable and practical for educators to implement in their teaching practices. To enhance model interpretability, educational institutions should employ interpretable machine learning techniques or develop simplified models that balance accuracy and comprehensibility. Techniques such as decision trees~\cite{kotsiantis2013decision} can be more transparent than complex neural networks. Visualization tools such as Tableau or Plotly can aid in making complex model outputs more understandable for educators. Moreover, it is crucial to translate model outputs into actionable insights through user-friendly dashboards and reporting tools. Collaborating with educators to identify key metrics and indicators ensures that the insights are meaningful and practical.

\subsection{Fairness and Privacy}
The collection and analysis of educational data raise significant privacy and ethical issues. Ensuring student privacy and complying with data protection regulations are critical. There is also the ethical concern of how the data is used, particularly in predictive analytics, where there is a risk of profiling or bias. However, ensuring that EDM models do not perpetuate or amplify biases is a significant challenge.

Such issues might be addressed from the following aspects. Firstly, we could focus on developing and implementing fairness algorithms, such as equality of opportunity in supervised learning~\cite{hardt2016equality}, which can help to identify and rectify biases in predictive models. Additionally, research could aim to improve techniques for bias detection in training data and outcomes. Secondly, to protect students' privacy, the development and application of privacy-preserving data mining techniques should be a key focus.
To this end, we can adopt Differential Privacy Stochastic Gradient Descent (DP-SGD)~\cite{xie2021differential} or Private Aggregation of Teacher Ensembles (PATE)~\cite{papernot2016semi} to train models without directly accessing sensitive data. Lastly, educational institutions could explore the implementation and optimization of federated learning algorithm~\cite{li2020federated} in EDM. It enables model training on local data without having to share it with a central server, thereby protecting student information.

Overall, EDM holds transformative potential for enhancing educational practices and learning outcomes. However, realizing this purpose requires addressing several practical challenges. By implementing targeted strategies to overcome the above challenges, educational institutions can effectively leverage data mining techniques to offer more personalized and effective learning experiences.

\section{Future Directions}
Deep Learning has shown great promise in EDM. Successful techniques in EDM based on Deep Learning can provide valuable insights to improve teaching, learning, and assessment. Here are several potential insights into successful techniques for EDM using Deep Learning:

\subsection{Learning Analysis and Intervention}
Existing EDM methods often involve offline analysis. Future work could investigate the use of real-time learning algorithms, \eg deep reinforcement learning~\cite{lin2023survey} or real-time recurrent learning algorithms~\cite{menick2020practical}, to offer timely insights and interventions for a more responsive educational environment.
In addition, we can combine Multi-Task Learning~\cite{zhang2021survey} with Attention mechanisms to analyze student interaction data from learning management systems or other educational platforms, providing knowledge tracing and option tracing~\cite{an2022no} into engagement and learning patterns. Multi-Task Learning can also be used to identify learners at risk of struggling academically and dropping out, thus allowing for early interventions and support.

\subsection{Social Network Analysis and Collaboration}
By learning directly from the graph structure and node features, GCNs can capture complex patterns in social networks, helping to analyze social networks within educational settings, revealing patterns of collaboration and communication among students.
These insights can help educators design group activities and assignments more effectively or identify students who may benefit from additional support or social engagement opportunities. Besides, future work should encourage collaborations between computer scientists, educators, and psychologists. Although this direction does not focus on a specific algorithm, it emphasizes the importance of interdisciplinary knowledge in refining existing algorithms or developing new ones for EDM. For example, to make cross-domain recommendations for users in the educational environment, we can use the preference-aware Graph Attention Network~\cite{li2023preference}, which leverages collaborative KG to capture user preferences within-domain and across-domain.

\subsection{Explainable AI in EDM}
Given the ‘black box’ problem associated with Deep Learning models, efforts should be made to create more transparent and interpretable models. Especially in EDM, making these models explainable and transparent becomes increasingly important, since education usually places great emphasis on the scientific nature and causal relationships of things. Future research could aim to develop or improve methods for generating understandable explanations of model predictions, \eg Deep Learning Important FeaTures DeepLIFT)~\cite{shrikumar2017learning} for the interpretable models used in student performance prediction, and Local Interpretable Model-Agnostic Explanations (LIME)~\cite{zhao2021baylime} or KGs for explainable recommendation systems.
Moreover, for applications that involve sequential data, such as studying student interaction with a learning management system over time, models like LSTM or GRU could be enhanced with explainability features. To this end, sequence interpretation methods, \eg Layer-wise Relevance Propagation (LRP)~\cite{montavon2019layer}, can be used to explain sequential learning patterns.

\subsection{Large Language Models for Education}
Large Language Models (LLMs)~\cite{min2021recent} are transforming many fields, including EDM. Their capacity to understand, generate, and complete texts makes them a valuable tool in education. For instance, LLMs could be used to generate personalized educational content tailored to each learner's needs, interests, and proficiency level. To this end, we could focus on in-context learning or task-specific prompt methods for LLMs, such as GPT-4, to achieve better performance in such tasks. Besides, LLMs could help create advanced Intelligent Tutoring Systems (ITSs)~\cite{mousavinasab2021intelligent} that can understand and respond to student queries in a contextually appropriate manner. Future work should integrate LLMs into existing ITSs and examine their impact on student learning outcomes.

It would be interesting that GPT-based architectures~\cite{zhang2023complete, sanderson2023gpt} are employed to automatically score student essays or assess written responses, which can reduce the workload for educators and provide consistent evaluation. Such architectures are also competent to analyze and understand student language use, enabling educators to identify areas where students even teachers struggle with understanding or communication.

The LLM agent~\cite{Liao23} represents a groundbreaking development in the realm of education technology. The LLM agent excels in generating high-quality educational content. Thus, learners can interact with the agent in a conversational manner, enabling a more personalized and adaptive learning experience tailored to individual needs. Moreover, the LLM agent promotes inclusivity in education by offering multilingual support. It can seamlessly communicate in various languages, breaking down language barriers and providing access to educational resources for a global audience. 

\subsection{Multimodal Learning Analytics}
Many current EDM methods primarily rely on structured data. However, educational experiences produce a wealth of unstructured and semi-structured data (\eg image, audio, video, and even biometric data), which provide a more comprehensive understanding of student learning experiences. Deep Multimodal Learning algorithms~\cite{ramachandram2017deep}, which combine CNNs for image/video data, RNNs or LSTMs for temporal data, and Transformers for textual data, could be used to leverage this wealth of information.
These insights can inform the design of multimodal learning environments and interventions, accommodating diverse learning styles and preferences.

Based on the above representation of multimodal learning, multimodal affective computing and emotion recognition may be a promising line in EDM. 
To this end, we need well-designed models to analyze student facial expressions, speech, or physiological signals to infer emotional states and engagement during learning activities. In this way, they help educators adapt their teaching strategies to better meet students' emotional needs and improve overall learning experiences. For instance, hybrid contrastive learning~\cite{mai2022hybrid} can be employed for multimodal sentiment analysis, in which semi-contrastive learning and intra-/inter-modal contrastive learning learn multiple relationships from cross-modal interactions. 

In summary, while Deep Learning has already demonstrated its potential in EDM, there are quantities of exciting opportunities for further exploration and innovation. By focusing on the future directions highlighted above, we hope to promote significant progress in the field, contributing to the transformation of educational practices and the improvement of educational experiences.

\section{Conclusion}
Deep Learning algorithms have been widely applied in various fields. They have shown great potential in EDM to assist in improving the quality of modern education. In this survey, we first offer an extensive outline of the current state-of-the-art in Deep Learning-based EDM, highlighting three categories of Deep Learning (\ie unsupervised learning, supervised learning, and reinforcement learning) applied to four main educational scenarios. Additionally, we draw designs for knowledge tracing schema, student behavior detection's schema, and personalized recommendation framework, to demonstrate their principles intuitively. Secondly, a thorough overview of public datasets and processing tools for EDM is elaborated. Thirdly, We conduct a detailed analysis of the practical challenges in EDM and present targeted solutions. Lastly, to provide new opportunities for innovation and improvement in this area, we put forward some promising future directions. This survey aims to inspire further research, collaboration, and progress toward broadening the application scope of EDM with Deep Learning.

\section{Acknowledgments}
This work is supported by the National Natural Science Foundation of China (No. 61977055).

\printcredits
\bibliographystyle{cas-model2-names}

\bibliography{cas-refs}


\end{document}